\documentclass[sigconf, screen]{acmart} % , authorversion, nonacm

\usepackage{multirow}
\usepackage{balance}
\usepackage[table]{xcolor}
\usepackage[capitalize]{cleveref}
\Crefname{section}{Section}{Sections}
\crefname{section}{Sec.}{Secs.}
\Crefname{table}{Table}{Tables}
\crefname{table}{Tab.}{Tabs.}
\Crefname{figure}{Figure}{Figures}
\crefname{figure}{Fig.}{Figs.}
\Crefname{table}{Table}{Tables}
\crefname{table}{Tab.}{Tabs.}
\Crefname{equation}{Equation}{Equations}
\crefname{equation}{Eq.}{Eqs.}
\Crefname{algorithm}{Algorithm}{Algorithms}
\crefname{algorithm}{Algo.}{Algos.}
\Crefname{theorem}{Theorem}{Theorems}
\crefname{theorem}{Thm.}{Thms.}

\newcommand{\eg}{\textit{e.g.}}
\newcommand{\ie}{\textit{i.e.}}
\newtheorem{theorem}{Theorem}
\newcommand{\gcell}{\cellcolor{gray!15}}

\renewcommand\footnotetextcopyrightpermission[1]{} %remove copyright
\AtBeginDocument{%
  }

\copyrightyear{2026}
\acmYear{2026}
\setcopyright{cc}
\setcctype{by}
\acmConference[MM '26]{Proceedings of the 34th ACM International Conference on Multimedia}{November 10--14, 2026}{Rio de Janeiro, Brazil}
\acmBooktitle{Proceedings of the 34th ACM International Conference on Multimedia (MM '26), November 10--14, 2026, Rio de Janeiro, Brazil}
\acmDOI{10.1145/3767308.3835280}
\acmISBN{979-8-4007-2213-4/2026/11}

\acmSubmissionID{2165}

\begin{document}

\title[SteerFace: Debiasing Synthetic Face Generation via Adaptive Residue Perturbation]{SteerFace: Debiasing Synthetic Face Generation \\ via Adaptive Residue Perturbation}

\author{Yuxi Mi}
\affiliation{\institution{Fudan University}  \city{Shanghai}  \country{China}}
\email{yxmi20@fudan.edu.cn}

\author{Qiuyang Yuan}
\affiliation{\institution{Fudan University}  \city{Shanghai}  \country{China}}
\email{qyyuan23@m.fudan.edu.cn}

\author{Jianqing Xu}
\affiliation{\institution{Youtu Lab, Tencent} \city{Shanghai}  \country{China}}
\email{joejqxu@tencent.com}

\author{Yichun Zhou}
\affiliation{\institution{Fudan University}  \city{Shanghai}  \country{China}}
\email{yichunzhou25@m.fudan.edu.cn}

\author{Xuan Zhao}
\affiliation{\institution{Fudan University}  \city{Shanghai}  \country{China}}
\email{xzhao23@m.fudan.edu.cn}

\author{Jun Wang}
\affiliation{\institution{WeChat Pay Lab33, Tencent} \city{Shenzhen}  \country{China}}
\email{earljwang@tencent.com}

\author{Rizen Guo}
\affiliation{\institution{WeChat Pay Lab33, Tencent} \city{Shenzhen}  \country{China}}
\email{rizenguo@tencent.com}

\author{Shuigeng Zhou}\correspondingauthor%\authornote{Corresponding author.}
\affiliation{\institution{Fudan University}  \city{Shanghai}  \country{China}}
\email{sgzhou@fudan.edu.cn}

\renewcommand{\shortauthors}{Yuxi Mi et al.}

\begin{abstract}

The shortage of legally compliant data for face recognition training has sparked growing interest in using synthetic data as an alternative. While recent diffusion-based methods enable the generation of photorealistic face images with strong identity adherence and data diversity, their downstream recognition performance still exhibits a significant synthetic-real gap. This paper identifies visual tendency as a previously underexplored limitation, whereby synthetic data exhibit an unrealistic prevalence of visual attributes and thus deviate from the real-data distribution. Visual tendency can be attributed to the generator's conditioning on identity embeddings, through which co-occurring residual visual cues are unintentionally absorbed into learned identity semantics. To discourage the generator from exploiting such visual cues, this paper proposes SteerFace, a simple and efficient training framework that perturbs identity embeddings by steering them toward random orthogonal directions on the embedding hypersphere. The perturbation serves as an identity-preserving regularizer that penalizes the generator's reliance on non-identity components, as supported by theoretical analysis. This paper further introduces an adaptive strategy that learns perturbation strengths with both sample-wise preference and favorable overall statistics. Experiments show that SteerFace mitigates visual tendency, improves downstream recognition performance, and generalizes across settings.
% Extensive experiments show that SteerFace effectively mitigates visual tendency, outperforms prior methods in downstream face recognition, and generalizes well across different training datasets and generation pipelines.

\end{abstract}






    \maketitle

\begin{figure}[tbp]
    \centering
    \includegraphics[width=\linewidth]{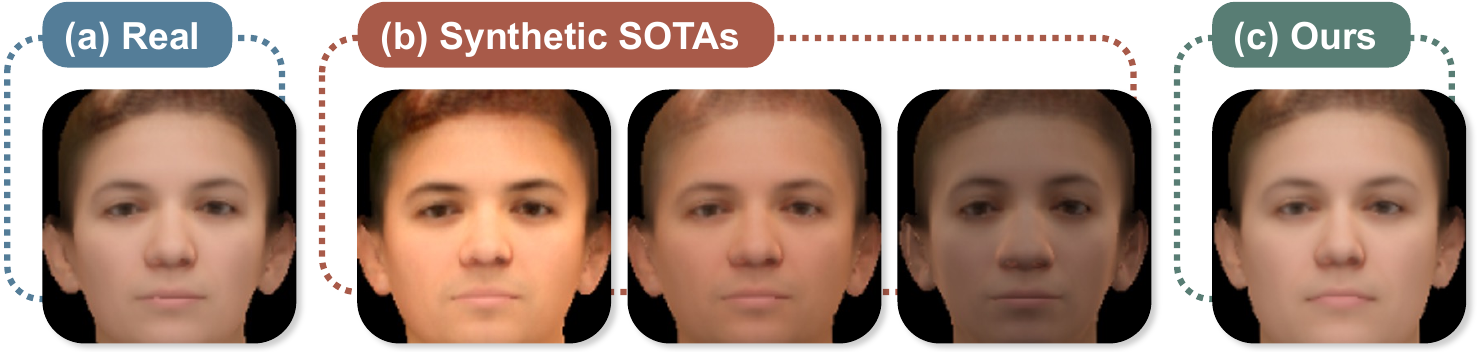}
    \caption{Visual tendency in synthetic face generation, illustrated by 3DMM rendering of the dataset-wise average face~\cite{feng2021learning}. Compared with (a) real data~\cite{yi2014learning}, (b) synthetic SOTAs~\cite{lin2025uiface,mi2025data,bae2023digiface} exhibit varying degrees of visual leaning, indicating a distribution bias. (c) SteerFace mitigates this bias.}

    % for internal review
%     \caption{Visual tendency in synthetic face generation, illustrated by 3DMM rendering of the dataset-wise average face. Compared with (a) real data, (b) synthetic SOTAs exhibit varying degrees of visual leaning, indicating a distribution bias. (c) SteerFace mitigates this bias. NOTE: Face images are from CASIA~\cite{feng2021learning}, UIFace, MorphFace, and DigiFace~\cite{lin2025uiface,mi2025data,bae2023digiface}.}
    
    \label{fig:paradigm}
\end{figure}

\section{Introduction}
\label{sec:intro}

% TODO: 
% 1. 从3DMM观察到有偏到归因于ID，这里可能有个gap，需要一句话解释，或ref到后文分析
% ~~2. 比context dropout好在哪，可能需要说一句~~ Done

Face recognition (FR) is among the most prominent applications of computer vision. Its success relies on large-scale training datasets of face images from tens of thousands of individuals~\cite{zhu2021webface260m,kemelmacher2016megaface}. However, the legitimate acquisition of face data remains a crucial challenge: most existing datasets were web-crawled without informed consent~\cite{lopez2022legal}, which has raised ethical concerns and led to the withdrawal of several well-known datasets~\cite{cao2018vggface2,guo2016ms}. This shortage of legally compliant data has sparked interest in synthetic data of ``virtual'' persons as alternatives for training FR models~\cite{frcsyn2025,deandres2024frcsyn}.

% feature diversity?
Ideally, synthetic data should mimic real data closely enough that FR models trained on them generalize effectively to real persons. Prior studies typically assess this goal through two proxy criteria: \textit{identity adherence}~\cite{boutros2023idiff,kim2023dcface}, which requires each synthetic image to faithfully preserve an assigned virtual identity; and \textit{data diversity}~\cite{lin2025uiface,papantoniou2024arc2face,li2024id,mi2025data,boutros2026idperturb}, which requires the dataset as a whole to cover broad real-world variations. With recent advances in diffusion models (DMs)~\cite{ho2020denoising,song2020denoising,rombach2022high}, state-of-the-art (SOTA) methods have made substantial progress on both criteria, generating photorealistic face images with strong identity consistency and rich variation. Nevertheless, their FR performance still exhibits a significant synthetic-real gap~\cite{frcsyn2025,deandres2024frcsyn}. This discrepancy suggests that the prevailing criteria, while important, may not fully capture the aspects of synthetic data quality that matter for FR.

To probe what may still be missing, this paper revisits synthetic data from a different dataset-level perspective. While data diversity measures the richness of variation, it remains unclear whether such variations appear with realistic prevalence. To examine this point, given a dataset, this paper employs a 3D morphable model (3DMM)~\cite{feng2021learning} to parameterize each face image by a set of visual attributes, including shape, expression, texture, and illumination. These attributes are averaged over the dataset and rendered into an ``average face'', which represents the dataset's overall visual tendency. \Cref{fig:paradigm}(a-b) compares the average faces of synthetic SOTAs with that of real-world ground truth, and reveals clear differences, \eg, in color tone. This suggests that a synthetic dataset can be diverse yet still exhibit an overall visual leaning, yielding a \textit{distribution bias} that weakens FR performance.

This paper traces the bias largely to how DMs learn identity adherence. As a \textit{de facto} training practice, SOTAs pair each image with its identity embedding, commonly extracted via an off-the-shelf FR model, so that the DM learns to reproduce the image of the prescribed person. While such embeddings are identity-discriminative, prior studies have shown that they are also entangled with rich visual cues~\cite{deng2019arcface,huang2020curricularface,zhong2024slerpface}. As a result, the DM may unintentionally absorb these cues as part of identity. Once learned in this way, the condition no longer specifies only who the person is, but also implicitly binds that identity to biased visual preferences.

To address this bias, this paper proposes a simple and efficient training scheme, \textit{SteerFace}, which perturbs identity during training by geometrically steering it in the embedding space. The perturbation regularizes the DM's reliance on co-occurring residual visual cues, as shown in~\cref{fig:paradigm}(c). Importantly, unlike common regularization techniques such as context dropout~\cite{ho2022classifier,boutros2023idiff}, it acts \textit{selectively} on non-essential cues, thereby reducing distribution bias while better maintaining identity adherence. It also improves data diversity by allowing visual attributes to vary more freely, rather than being bound to identity. This paper further proposes an adaptive strategy that learns an improved sample-aware allocation of perturbation intensity. Experiments show that SteerFace is broadly effective across different training datasets and generation pipelines, and achieves downstream FR performance superior to that of all SOTAs. \footnote{Sample face images shown in this paper are drawn from CASIA-WebFace~\cite{yi2014learning}, prior synthetic SOTAs~\cite{boutros2022sface,bae2023digiface,boutros2023idiff,kim2023dcface,lin2025uiface,mi2025data}, and the proposed SteerFace. All images are used in accordance with the applicable data-use terms.}

% The perturbation regularizes the DM's reliance on co-occurring visual cues, as theoretically shown, and enables SteerFace to reduce distribution bias without sacrificing identity adherence.

Overall, this paper presents three main contributions:
% \begin{itemize}
%     \item We identify the biased visual tendency arising from identity-conditioned learning as a major challenge in synthetic data quality for FR.
%     \item We propose a training scheme, SteerFace, that reduces this bias by geometrically and adaptively perturbing identity during training.
%     \item We present extensive analyses showing that our generated data achieve superior and more generalizable performance than those of SOTAs.
% \end{itemize}
\begin{itemize}
    \item We identify biased visual tendency in identity-conditioned face generation as a major challenge for synthetic FR data.
    \item We propose SteerFace, which reduces this bias by geometrically and adaptively perturbing identity during training.
    \item We present extensive analyses showing that SteerFace outperforms SOTAs in both generation quality and downstream FR performance.
\end{itemize}

\section{Related Work}
\label{sec:rw}

% --- update: lessly relevant citations can be trimmed a little bit ---

% \subsection{Face Recognition}

% The success of modern FR is built on the progress of robust feature extractors~\cite{he2016deep}, discriminative objectives~\cite{boutros2022elastic, deng2019arcface, huang2020curricularface, wang2018cosface, kim2022adaface}, and large-scale training datasets~\cite{cao2018vggface2, zhu2021webface260m, guo2016ms, kemelmacher2016megaface}. However, acquiring diverse and legally compliant datasets remains a crucial challenge~\cite{lopez2022legal}, and ethical disputes have led to the withdrawal of several major datasets~\cite{guo2016ms,cao2018vggface2}. This has sparked interest in using synthetic data as an alternative source for training FR models~\cite{frcsyn2025,deandres2024frcsyn}.

\subsection{Face Image Synthesis}

Generating high-quality face images is a long-standing challenge. Early works primarily employ 3D graphics~\cite{deng2018uv,geng20193d,piao2019semi,xu2024chain} or generative adversarial networks (GANs)~\cite{karras2019style,medin2022most,nguyen2019hologan}. The recent rise of DMs~\cite{ho2020denoising,song2020denoising,rombach2022high} has further enabled photorealistic face generation and plausible personalization to specific subjects by conditioning the DM on reference images~\cite{ruiz2023dreambooth,gal2022image,ding2023diffusionrig}, identity-descriptive features~\cite{xiao2024fastcomposer,li2024photomaker,yuan2023inserting,wang2024stableidentity,valevski2023face0,peng2024portraitbooth}, or textual prompts~\cite{gal2023encoder,zhou2023enhancing} of the person. Though these methods have shown success in tasks such as video generation~\cite{guo2024liveportrait,zhong2025anytalker,xu2025clgc,xu2026disco,xu2026smrabooth} and privacy preservation~\cite{mi2024privacy,zhong2024slerpface}, they are not readily suitable for FR training, due to insufficient identity retention~\cite{boutros2023idiff} and limited diversity in generation at scale~\cite{frcsyn2025}.

\subsection{Face Synthesis for Recognition} % Identity-Preserving Face Synthesis
\label{subsec:rw-ipfs-sota}

Synthetic data for FR training requires \textit{scalable} generation of face images that each accurately adhere to virtual identities and together exhibit rich variation that mimics the diversity of faces in the wild. To retain identity, early GAN-based works such as SynFace~\cite{qiu2021synface} manipulate the identity latent space; SFace~\cite{boutros2022sface}, SFace2~\cite{boutros2024sface2}, IDNet~\cite{kolf2023identity}, and ExFaceGAN~\cite{boutros2023exfacegan} adopt StyleGAN~\cite{karras2019style} for identity-conditioned generation. Among DM-based methods, IDiff-Face~\cite{boutros2023idiff} conditions DM on identity embeddings extracted from a pretrained FR model, which has since become a \textit{de facto} practice. To enrich diversity, DigiFace~\cite{bae2023digiface} utilizes 3D priors to produce distinctive yet less realistic faces, IDiff-Face~\cite{boutros2023idiff} and UIFace~\cite{lin2025uiface} partially unconstrain identity conditions, DCFace~\cite{kim2023dcface} uses feature banks from auxiliary images, Arc2Face~\cite{papantoniou2024arc2face} inherits the knowledge of pre-trained Stable Diffusion~\cite{rombach2022high}, CemiFace~\cite{sun2024cemiface} mines semi-hard samples, ID3~\cite{li2024id} injects explicit facial attributes, MorphFace~\cite{mi2025data} uses 3DMM renderings as contexts, USynthFace~\cite{boutros2023unsupervised} learns from unlabeled synthetic faces, Vec2Face~\cite{wu2025vec2face} samples and perturbs identity vectors, HyperFace~\cite{shahreza2025hyperface} optimizes hyperspherical identity packing, NegFaceDiff~\cite{caldeira2025negfacediff} introduces negative identity conditions, and IDPerturb~\cite{boutros2026idperturb} employs inference-time perturbations on identity conditions. While these works successfully improve data diversity, their generation commonly exhibits a distribution bias in visual tendency from real data, which hinders FR performance. Addressing this bias is the primary goal of this paper.

\section{Methodology}
\label{sec:method}

\subsection{Preliminary}
\label{sec:method-preliminary}

The latent diffusion model (LDM)~\cite{rombach2022high} allows the generation of photorealistic face images. It is trained under the DDPM~\cite{ho2020denoising} framework to predict the noise added to a latent code $\mathbf{z}\in\mathbb{R}^{c\times h \times w}$. Given an input image $\mathbf{x}\in\mathbb{R}^{C\times H \times W}$, it is mapped into the latent code via a pre-trained encoder $\mathcal{\phi}_e$ as $\mathbf{z}=\mathcal{\phi}_e(\mathbf{x})$, and is then perturbed by Gaussian noise $\boldsymbol{\epsilon}\sim \mathcal{N}(0,\mathbf{1})$ over $T$ discrete steps, % toward a nearly Gaussian distribution,
\begin{figure}[tbp]
    \centering
    \includegraphics[width=\linewidth]{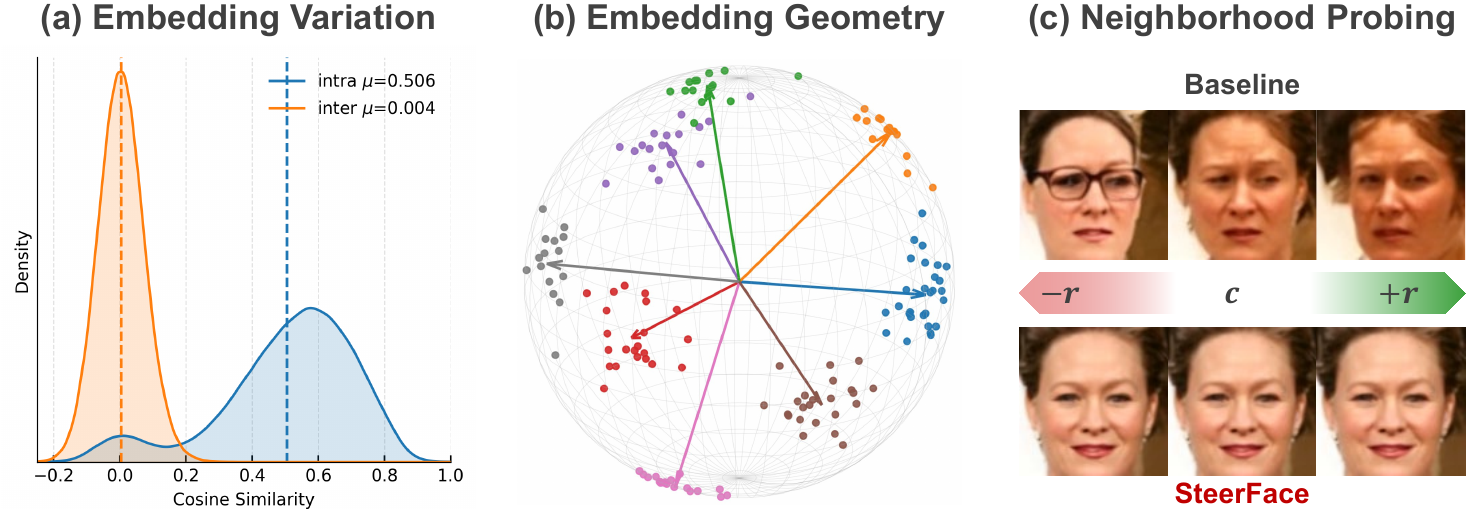}
    \caption{
    %Cause and effect of visual tendency. (a) Embeddings exhibit image-wise variation due to co-occurring visual cues, reflected by widely varying intra-class similarity. (b) Geometrically, embeddings (marked $\bullet$) are distributed around the ideal identity (marked $\uparrow$) on the hypersphere. (c) Under local identity perturbation, a baseline generator~\cite{boutros2023idiff} changes appearance due to entangled visual cues; SteerFace largely preserves appearance, indicating mitigated visual tendency.
    Cause and effect of visual tendency. (a) Embeddings vary across images due to co-occurring visual cues, reflected by widely varying intra-class similarity. (b) They ($\bullet$) lie around the ideal identity ($\uparrow$) on the hypersphere. (c) Under local identity perturbation, a baseline generator~\cite{boutros2023idiff} changes appearance, whereas SteerFace largely preserves it, indicating reduced visual tendency.
    % for internal review
    % NOTE: All face images shown are generated via~\cite{boutros2023idiff} based on synthetic virtual subjects~\cite{lin2025uiface}.
    }
    \label{fig:tendency}
\end{figure}

\begin{figure*}[tbp]
    \centering
    \includegraphics[width=\linewidth]{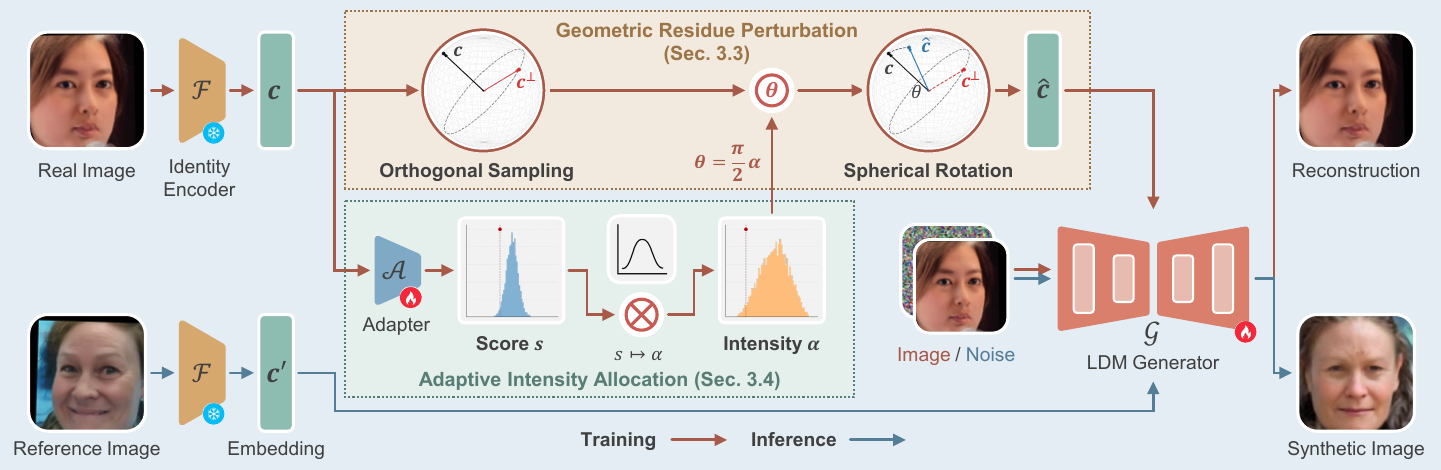}
    \caption{Pipeline of SteerFace. To mitigate visual tendency in synthetic face generation, during training, we perturb the identity embedding so that the generator is discouraged from exploiting visual cues. The perturbation is instantiated by spherically rotating the embedding toward a random orthogonal direction. The perturbation intensity is adaptively allocated to be both statistically favorable and sample-aware; see also~\cref{fig:mapping}. At inference time, generation follows the standard pipeline. % All face images shown are from synthetic virtual subjects~\cite{lin2025uiface}.
    }
    \label{fig:pipeline}
\end{figure*}

\begin{equation}
    \label{eq:ldm-noise}
    \mathbf{z}_{t} = \sqrt{\bar{\alpha}_{t}}\mathbf{z}_{0} + \sqrt{1-\bar{\alpha}_{t}}\boldsymbol{\epsilon},
\end{equation}

\noindent where $\mathbf{z}_{0}$ denotes the clean latent code, $\alpha_i$ is derived from a linearly scheduled variance term, and $\bar{\alpha}_t=\prod_{i=1}^{t}{\alpha_i}$. In the denoising process, the model predicts the noise via an estimator $\boldsymbol{\epsilon}_{\theta}$ (\eg, a U-Net~\cite{ronneberger2015u}) and parameterizes the reverse transition from $\mathbf{z}_t$ to $\mathbf{z}_{t-1}$,

% the model attempts to recover $\mathbf{z}_{t-1}$ iteratively through the transition modeled by a learnable noise estimator $\boldsymbol{\epsilon}_{\theta}$ (\eg, a U-Net~\cite{ronneberger2015u}),

\begin{equation}
    \label{eq:ldm-denoise}
    \mathbf{z}_{t-1} = \frac{1}{\sqrt{\alpha_t}} \left( \mathbf{z}_t - \frac{1 - \alpha_t}{\sqrt{1 - \bar{\alpha}_t}} \boldsymbol{\epsilon}_{\theta}(\mathbf{z}_t, t, \mathbf{c}) \right) + \sqrt{1-\alpha_t}\boldsymbol{\eta},%\boldsymbol{\eta}\sim \mathcal{N}(0,1),
\end{equation}

\noindent where $\boldsymbol{\eta}\sim \mathcal{N}(0,1)$. $\mathbf{z}_0$ is decoded into the image space as $\mathbf{x} = \mathcal{\phi}_d(\mathbf{z}_0)$. $\mathbf{c}\in \mathbb{R}^d$ is a context embedding that conditions the recovery, injected into the noise estimator through cross-attention. We hereafter omit the latent encoding $\mathbf{x}\mapsto\mathbf{z}$ for notational simplicity. % In identity-conditioned LDMs~\cite{boutros2023idiff,lin2025uiface,mi2025data,boutros2026idperturb}, the context $\mathbf{c}$ is typically an identity embedding extracted via a pre-trained FR model and projected into the noise estimator through cross-attention.

\subsection{Visual Tendency in Synthetic Data} 
\label{subsec:method-motivation}

We begin by explaining why synthetic data can exhibit visual tendency. The general goal of synthetic face generation for FR is to create a scalable set of real-looking face images $\mathbf{X}=\{\mathbf{x}_1,\dots,\mathbf{x}_n\}$ that adhere to many virtual identities $V=\{v_1,\dots,v_m\}$, thereby serving as an alternative for FR training. A common paradigm in recent studies~\cite{boutros2023idiff,kim2023dcface,lin2025uiface,mi2025data,boutros2026idperturb,li2024id} is to train an LDM generator $\mathcal{G}$ on legally compliant real data (\eg, CASIA~\cite{yi2014learning}). Identity adherence is encouraged by training the generator to reconstruct a face image $\mathbf{x}$ conditioned on its identity embedding $\mathbf{c}$ (\ie, $\mathcal{G}:(\mathbf{x},\mathbf{c})\mapsto \mathbf{x}$),

\begin{equation}
    \label{eq:ldm-goal}
    \mathcal{L}_{\mathrm{LDM}} = \mathbb{E}_{\mathbf{x},t,\boldsymbol{\epsilon}}{\left[||\boldsymbol{\epsilon}_{\theta}(\mathbf{x}_t,t,\mathbf{c})-\boldsymbol{\epsilon}||_2^2\right]},\quad 0<t\leq T.
\end{equation}

Ideally, the generator would learn from a ``pure'' embedding $\mathbf{c}_v$ that encodes only identity. However, such an embedding is hard to obtain, as face images do not present identity in isolation. In practice, SOTAs use embeddings extracted via an off-the-shelf FR model $\mathbf{c}=\mathcal{F}(\mathbf{x})$, which are identity-discriminative unit vectors by design. This is a compromise, as such embeddings also retain information about co-occurring visual attributes entangled in a particular image~\cite{deng2019arcface,wang2018cosface}, \eg, shape, expression, texture, and illumination, thus exhibiting image-wise variation. \Cref{fig:tendency}(a) shows the pairwise intra-class similarity of embeddings from CASIA~\cite{yi2014learning}, which spans a fairly wide range rather than collapsing to a single value. Geometrically, embeddings from different images of the same person form a local neighborhood on an $\mathbb{S}^{d-1}$ unit hypersphere around the hypothetical pure identity $\mathbf{c}_v$, as shown in~\cref{fig:tendency}(b). Formally, we can write $\mathbf{c}$ as the composition of $\mathbf{c}_v$ and a residual term $\mathbf{r}$:

\begin{equation}
    \label{eq:residue}
    \mathbf{c} = \mathrm{norm}(\mathbf{c}_v + \mathbf{r}), \quad \mathbf{r} \perp \mathbf{c}_v.
\end{equation}

Importantly, the residue is problematic for identity-conditioned generation. During training, any component of $\mathbf{c}$ that is predictive of the target face appearance, whether identity-related or not, can help minimize the reconstruction objective. In particular, conditioning on $\mathbf{r}$ in addition to $\mathbf{c}_v$ cannot make the optimal reconstruction worse:

\begin{equation}
    \label{eq:shortcut-residue}
    \min_\theta\mathbb{E}{\left[||\boldsymbol{\epsilon}_{\theta}(\mathbf{x}_t,t,\mathbf{c}_v)-\boldsymbol{\epsilon}||_2^2\right]} \geq \min_\theta\mathbb{E}{\left[||\boldsymbol{\epsilon}_{\theta}(\mathbf{x}_t,t,\mathbf{c}_v,\mathbf{r})-\boldsymbol{\epsilon}||_2^2\right]},
\end{equation}

\noindent with strict improvement whenever $\mathbf{r}$ is predictive of the denoising target. The generator is therefore encouraged to exploit not only identity but also the visual cues carried by the residue as shortcuts for reconstruction. In this process, it learns a biased coupling between identity and visual cues.

At inference, the trained generator synthesizes new faces $\mathbf{x}'$ from Gaussian noise and a virtual identity embedding $\mathbf{c}'$, obtained either from a prior distribution~\cite{li2024id} or from the reference image of a virtual person~\cite{boutros2023idiff,lin2025uiface,mi2025data}, \ie, $\mathcal{G}:(\mathcal{N}(0,\mathbf{1}),\mathbf{c}')\mapsto \mathbf{x}'$. Under an ideal training outcome, the embedding should control only identity, while visual attributes vary freely according to the overall learned data distribution. However, the induced coupling causes the generator to also rely on the residue in $\mathbf{c}'$, thereby giving rise to certain visual tendency. To illustrate this effect, we fix the Gaussian noise and condition a SOTA generator~\cite{boutros2023idiff} on a specific $\mathbf{c}'$, then perturb $\mathbf{c}'$ within its local neighborhood to mimic residual variation. As shown in~\cref{fig:tendency}(c), this yields images of the same person with noticeable appearance changes, suggesting that the generator absorbs not only identity but also visual cues from $\mathbf{c}'$. Visual tendency has two adverse effects: it limits data diversity and, more importantly, introduces a distribution bias from real data.

\begin{figure}[tbp]
    \centering
    \includegraphics[width=\linewidth]{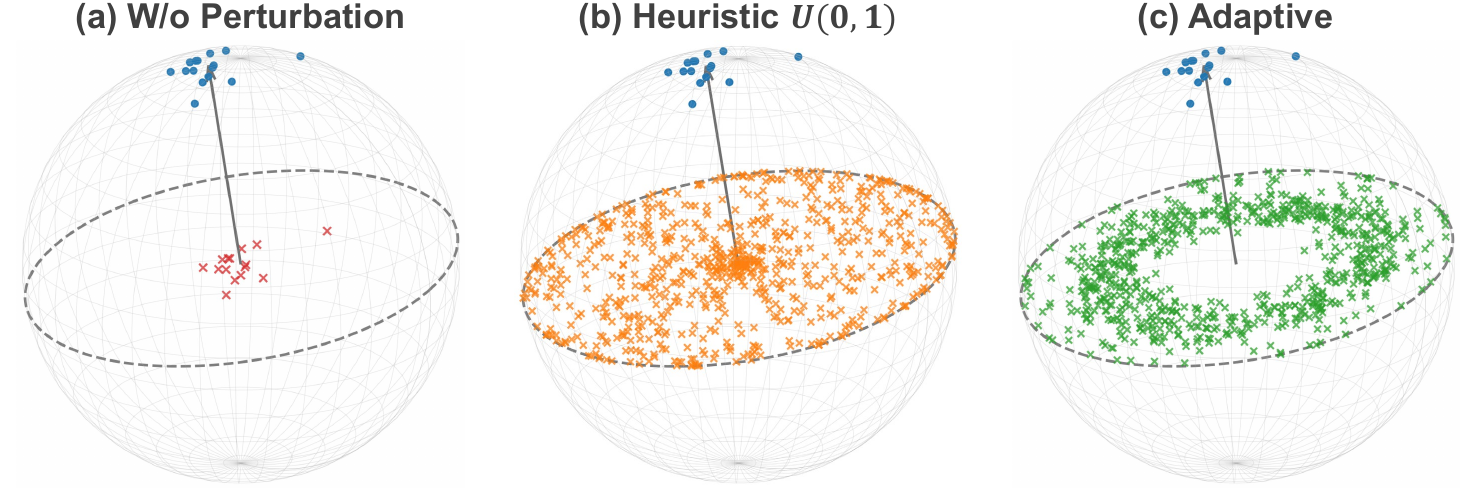}
    \caption{
    % Effect of SteerFace. We experimentally sample one CASIA identity, revisit each of its embedding (marked $\bullet$) 50 times, and plot the resulting residues ($\times$). (a) Without perturbation, residues are mapped to a deterministic location on the plane orthogonal to ideal identity ($\uparrow$), making them predictive and easy for the generator to exploit. (b) Heuristic perturbation turns residues into noise-like variations that spread across the plane at each visit, thereby regularizing generator learning. (c) Adaptive perturbation learns a spread with comparable identity retention but more favorable regularization, leading to better downstream FR performance.
    Effect of SteerFace. For one CASIA identity, we revisit each embedding ($\bullet$) 50 times and plot the resulting residues ($\times$). (a) Without perturbation, residues remain deterministic and thus exploitable. (b) Heuristic perturbation spreads noise-like residues over the plane orthogonal to the ideal identity ($\uparrow$), regularizing generator learning. (c) Adaptive perturbation achieves comparable identity retention with stronger regularization, improving downstream FR.
    }
    \label{fig:steer-effect}
\end{figure}

\subsection{Geometric Residue Perturbation}
\label{subsec:method-steer}

To mitigate visual tendency, we introduce \textit{SteerFace}, a simple and efficient generator training paradigm that discourages the learning of residue. \Cref{fig:pipeline} shows its overall pipeline. 

Our motivation stems from how the generator exploits the residue. We have shown in~\cref{eq:shortcut-residue} that the generator uses the residue as a shortcut when it is predictive for denoising, which happens because the residue is deterministically coupled with its source image and thus carries stable visual cues, as shown in~\cref{fig:steer-effect}(a). By contrast, consider an \textit{unpredictive} residue $\hat{\mathbf{r}}$, \ie, noise-like and (approximately) conditionally independent of the denoising target,

\begin{equation}
    \boldsymbol{\epsilon} \perp \hat{\mathbf{r}} \mid (\mathbf{x}_t,t,\mathbf{c}_v), \quad \mathbb{E}{[\boldsymbol{\epsilon} \mid \mathbf{x}_t,t, \mathbf{c}_v, \hat{\mathbf{r}}]} = \mathbb{E}{[\boldsymbol{\epsilon} \mid \mathbf{x}_t,t, \mathbf{c}_v]}.
\end{equation}

\noindent It then offers no gain for the $\mathcal{L}_{\mathrm{LDM}}$ objective, hence no incentive for the generator to rely on it. This motivates mitigating visual tendency by perturbing the deterministic residue $\mathbf{r}$ into a noise-like $\hat{\mathbf{r}}$ that changes at each visit, thereby destroying its predictability. 

Crucially, to maintain identity adherence, such perturbation should act on the residue $\mathbf{r}$ while preserving the identity component associated with $\mathbf{c}_v$. Geometrically, this corresponds to adding an identity-orthogonal random noisy component $\mathbf{n}\perp\mathbf{c}_v$ to the embedding $\mathbf{c}$, yielding $\hat{\mathbf{c}} = \mathrm{norm}(\mathbf{c}+\mathbf{n})$; thus, the residue is effectively corrupted as $\hat{\mathbf{r}}=\mathbf{r}+\mathbf{n}$ before normalization. Since the pure $\mathbf{c}_v$ is unavailable in practice, we instead use the good approximation that $\mathbf{c}_v$ is angularly close to $\mathbf{c}$, \ie, $\langle \mathbf{c}_v,\mathbf{c}\rangle \approx 1$, and choose $\mathbf{n}\perp\mathbf{c}$ accordingly. We further discuss the validity of this approximation in the supplementary material. % , as shown in {\color{blue}Fig.~4(a)}

We instantiate the perturbation by rotating the embedding toward a random orthogonal direction on the unit hypersphere $\mathbb{S}^{d-1}$. Concretely, each time a training face sample $\mathbf{x}$ is visited, we extract its embedding $\mathbf{c}$ and isotropically sample a unit vector $\mathbf{c}^{\perp}\perp\mathbf{c}$ by

\begin{equation}
    \label{eq:sample-orthogonal}
    \mathbf{c}^{\perp}=\mathrm{norm}\left(\mathbf{e}-(\mathbf{c}\cdot\mathbf{e})\mathbf{c}\right), \quad \mathbf{e}\sim\mathcal{N}(0,\mathbf{1}).
\end{equation}

\noindent Then, we rotate $\mathbf{c}$ toward $\mathbf{c}^{\perp}$ by an angle $\theta\in[0,\frac{\pi}{2})$. To enhance the unpredictability of the noisy residue, we introduce perturbations with varying intensities by sampling a different $\theta$ each time, parameterized as $\theta=\frac{\pi}{2}\alpha$, where $\alpha$ follows a prior distribution, \eg, $\alpha\sim U(0,1)$. We use $U(0,1)$ as a heuristic baseline, and revisit its extension later in~\cref{subsec:method-adaptive}. The rotation is given by

\begin{equation}
    \label{eq:rotation}
    \hat{\mathbf{c}}=\cos\theta\,\mathbf{c}+\sin\theta\,\mathbf{c}^{\perp}.
\end{equation}

\noindent From~\cref{eq:rotation}, the perturbed embedding is naturally decomposed into two parts: an identity component $\cos\theta\,\mathbf{c}$ and a noisy component $\sin\theta\,\mathbf{c}^{\perp}$, where the latter serves as $\mathbf{n}$. The generator is trained on perturbed embeddings as $\mathcal{G}:(\mathbf{x},\hat{\mathbf{c}})\mapsto \mathbf{x}$. At inference time, as the generator has learned to suppress residue cues, perturbation is no longer applied, and generation follows the standard conditional pipeline: $\mathcal{G}:(\mathcal{N}(0,\mathbf{1}),\mathbf{c}')\mapsto \mathbf{x}'$.

We show that such residue perturbation is both \textit{identity-preserving} and \textit{residue-regularizing}: During training, for embeddings $\{\mathbf{c}_1,\dots,\mathbf{c}_n\}$ of the same person, under the approximation $\langle \mathbf{c}_v,\mathbf{c}\rangle \approx 1$, the identity-bearing direction of each perturbed embedding remains aligned with $\mathbf{c}_v$, scaled only by $\cos\theta$. Since FR embeddings encode identity primarily through angular direction~\cite{deng2019arcface}, this preserves the overall identity structure while only mildly attenuating its training influence through scale. Meanwhile, their noisy components spread within the plane orthogonal to $\mathbf{c}_v$, as shown in~\cref{fig:steer-effect}(b), making the residue noise-like and less exploitable by the generator.

To validate the debiasing effect, we retrain the same SOTA generator~\cite{boutros2023idiff} from~\cref{subsec:method-motivation} on perturbed embeddings instead, and repeat the local-neighborhood probing. In~\cref{fig:tendency}(c), images of the same person remain largely unchanged despite residual variation, indicating that the generator has learned to disentangle identity from visual cues, thereby mitigating visual tendency. Results in~\cref{tab:exp-perturb} further show a 1.37\% FR accuracy gain with $U(0,1)$ perturbation, suggesting that it improves downstream FR significantly.

Formally, we show that the perturbation serves as an implicit regularizer for identity-conditioned generation:

\begin{theorem}
    \label{thm:regularizer}
    Minimizing the $\mathcal{L}_{\mathrm{LDM}}$ objective with perturbed embeddings $\hat{\mathbf{c}}$ defined in~\cref{eq:sample-orthogonal,eq:rotation} penalizes the generator's sensitivity to non-identity components. Regularization strength scales with $\mathbb{E}[\sin^2\theta]$ and expected identity retention is given by $\mathbb{E}[\cos\theta]$.
\end{theorem}

\noindent The proof is deferred to the supplementary material.

We distinguish SteerFace from two related lines of work: \textit{Context dropout}~\cite{ho2022classifier} regularizes LDM conditioning by suppressing all semantics indiscriminately, whereas SteerFace preserves identity in a structured way. Several SOTAs~\cite{li2024id,sun2024cemiface,boutros2026idperturb} have explored \textit{inference-time perturbation} to enrich data diversity. IDPerturb~\cite{boutros2026idperturb} is particularly related to ours in that it also exploits spherical rotation. By contrast, SteerFace applies perturbation during training to align synthetic and real data distributions. See~\cref{subsec:exp-perturb-sota} for discussion.

\subsection{Adaptive Intensity Allocation}
\label{subsec:method-adaptive}

We further investigate the allocation of perturbation intensity. Though the heuristic choice $\alpha \sim U(0,1)$ is empirically effective, it is neither \textit{statistically optimal} nor \textit{sample-aware}.

From a statistical perspective, \cref{thm:regularizer} shows that the regularizing effect of perturbation becomes stronger as $\mathbb{E}[\sin^2\theta]$ increases, yet at the cost of weakening identity retention, as the identity component $\mathbf{c}_v$ is attenuated by $\mathbb{E}[\cos\theta]$. In balancing the two, we compare different heuristic distributions later in~\cref{tab:exp-perturb}, and find that downstream FR generally favors stronger regularization. Meanwhile, prior studies~\cite{mi2025data,kim2023dcface,lin2025uiface} suggest that identity adherence need only be sufficient for the FR model to learn stable identity discrimination. This implies that, compared with $U(0,1)$, a distribution that preserves adequate identity retention while inducing stronger regularization may be more favorable for downstream FR. On the other hand, sampling $\alpha$ from $U(0,1)$ also falls short in perturbing all embeddings indiscriminately. In practice, embeddings differ in how strongly they are coupled with visual cues, and may therefore require different perturbation strengths.

\begin{figure}[tbp]
    \centering
    \includegraphics[width=\linewidth]{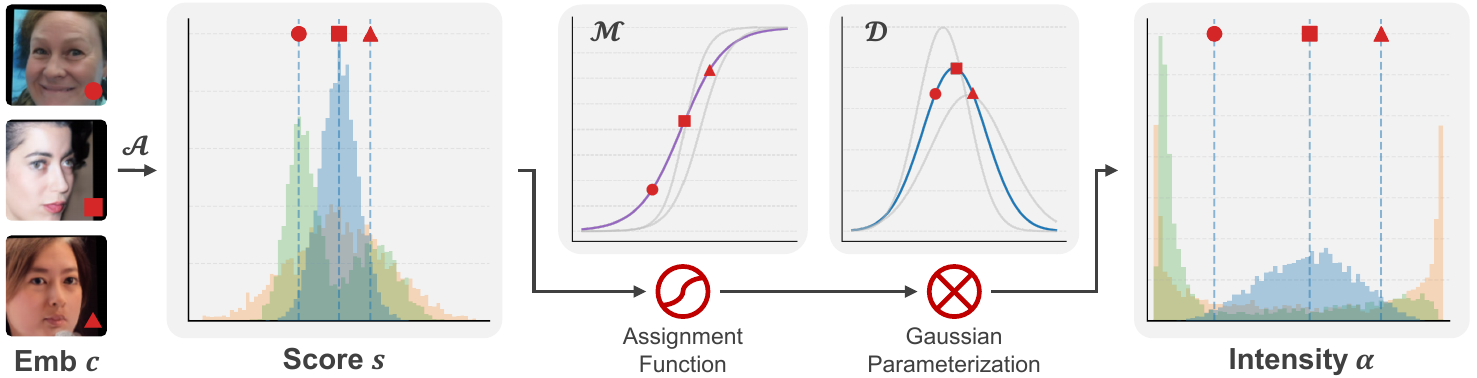}
    \caption{
    % Adaptive intensity allocation. A score $s$ of perturbation preference is assigned by adapter $\mathcal{A}$, and mapped to the final intensity $\alpha$ through assignment function $\mathcal{M}$ and Gaussian parameterization $\mathcal{D}$. The plots are experimentally derived. Notably, 1) $\mathcal{D}$ is adaptively parameterized; 2) the adapter can learn diverse score distributions, which, when composed with $\mathcal{M}$, induce expressive intensity allocations (in different colors); and 3) within the same distribution, score and intensity preserve monotonic order ($\bullet<\scalebox{0.8}{$\blacksquare$}<\blacktriangle$).
    Adaptive intensity allocation. Adapter $\mathcal{A}$ predicts a perturbation-preference score $s$, which is mapped to intensity $\alpha$ by assignment function $\mathcal{M}$ and adaptive Gaussian parameterization $\mathcal{D}$. The experimental plots show that diverse score distributions induce expressive intensity allocations (colors), while preserving monotonic score--intensity order within each distribution ($\bullet<\scalebox{0.8}{$\blacksquare$}<\blacktriangle$).
    % All shown faces depict synthetic identities~\cite{lin2025uiface}. 
    }
    \label{fig:mapping}
\end{figure}

To improve FR performance, we first seek an allocation of $\alpha$ whose overall statistics are more favorable than those of $U(0,1)$. However, directly learning a generic distribution is non-trivial, since the global constraints on identity retention and regularization are difficult to impose tractably. To simplify learning, we parameterize the allocation with a Gaussian family $\mathcal{N}(\mu,\sigma^2)$, whose overall shape can be controlled by only two parameters, $\mu$ and $\sigma$, under explicit constraints. Despite its simplicity, this parameterization remains sufficiently expressive in the final distribution shape when combined with the sample-aware assignment introduced next.% This choice is also sufficiently expressive in practice, as it will later be combined with a sample-aware assignment mechanism, and is therefore not overly restrictive on the final distribution shape.

Concretely, recall that the rotation angle is determined by the perturbation intensity as $\theta=\frac{\pi}{2}\alpha$. We parameterize the intensity as $\alpha = \mathcal{D}(z) = \mu + \sigma z$, where $z$ denotes a sample-wise coordinate to be assigned later. We then regularize the induced $\theta$ through two indicator statistics of identity retention and regularization, namely, $\mathbb{E}[\cos\theta]$ and $\mathbb{E}[\sin^2\theta]$. For identity retention, we want the learned allocation to remain comparable to $U(0,1)$. We therefore regularize $\mathbb{E}[\cos\theta]$ through the following consistency objective,

\begin{equation}
    \label{eq:loss-id}
    \mathcal{L}_{\mathrm{id}} =
    \left(
    \mathbb{E}[\cos\theta]
    -
    \mathbb{E}_{\alpha\sim U(0,1)}\left[\cos\left(\frac{\pi}{2}\alpha\right)\right]
    \right)^2.
\end{equation}

\noindent Meanwhile, to encourage stronger regularization, we impose a soft encouragement term that favors larger $\mathbb{E}[\sin^2\theta]$,

\begin{equation}
    \label{eq:loss-reg}
    \mathcal{L}_{\mathrm{reg}} = \frac{1}{1+k\,\mathbb{E}[\sin^2\theta]},
\end{equation}
\noindent where $k$ is a scaling coefficient, empirically set to 10. To improve upon $U(0,1)$ rather than learn from scratch, we further initialize $\mu$ and $\sigma$ to match those of $U(0,1)$, namely, $\mu_0=\frac{1}{2}, \sigma_0=\sqrt{1/12}$.

To enable sample-wise adaptation, we further train a lightweight MLP adapter $\mathcal{A}:\mathbf{c}\mapsto s$ jointly with the generator, which assigns each embedding $\mathbf{c}$ a scalar score $s\in\mathbb{R}$. The score indicates where the current embedding should be placed among all samples in terms of perturbation preference. To associate this score with the Gaussian family, we project it to a sample-wise coordinate through an assignment function $\mathcal{M}$. The function is monotonic, so that embeddings with stronger perturbation preference receive larger intensities, and nonlinear, so as to improve the expressiveness of the resulting allocation; see the supplementary material for details. The overall mapping from $\mathbf{c}$ to $\alpha$ is thus given by

\begin{equation}
    \label{eq:overall-mapping}
    \alpha=(\mathcal{D}\circ\mathcal{M}\circ\mathcal{A})(\mathbf{c}).
\end{equation}

Notably, though $\mathcal{D}$ itself follows a simple Gaussian parameterization, its composition with the adapter $\mathcal{A}$ and the nonlinear assignment $\mathcal{M}$ yields a sufficiently flexible allocation of $\alpha$, as illustrated in~\cref{fig:mapping} and later experimentally shown in~\cref{fig:perturb_dist}(b).

Combining the above with~\cref{eq:ldm-goal} yields the overall training objective of SteerFace under adaptive intensity allocation,

\begin{equation}
    \label{eq:loss-overall}
    \mathcal{L} = \mathcal{L}_{\mathrm{LDM}} + w_\mathrm{id}\,\mathcal{L}_{\mathrm{id}} + w_\mathrm{reg}\,\mathcal{L}_{\mathrm{reg}} + \mathcal{R}(\alpha),
\end{equation}

\noindent where $w_\mathrm{id}$ and $w_\mathrm{reg}$ are weighting hyper-parameters. $\mathcal{R}(\alpha)$ is a simple regularization term that prevents the dispersion of $\alpha$ from collapsing; see the supplementary material for details. \Cref{fig:steer-effect}(c) demonstrates the effect of adaptive allocation, where a more favorable distribution of perturbation intensity is learned in terms of identity retention and regularization. \Cref{tab:exp-perturb} further shows that it improves downstream FR by 0.51\% over the heuristic $U(0,1)$, and by 1.88\% over training without perturbation.

\begin{table*}[tbp]
\small
\centering
\caption{Comparison with SOTAs in terms of downstream FR accuracy. Results marked with an asterisk (*) are our reproductions. Bold and underlined numbers indicate the best and second-best results, respectively; the same applies hereafter.}
\label{tab:exp-fr-performance}

\begin{tabular}{llllcccccc}
\toprule
\textbf{Method} & \textbf{Venue} & \textbf{Generator} & \textbf{Scale (\#IDs × \#Imgs)} & \textbf{LFW} & \textbf{CFP-FP} & \textbf{AgeDB} & \textbf{CPLFW} & \textbf{CALFW} & \textbf{Average} \\
\midrule
CASIA~\cite{yi2014learning}           & (Real)       & /         & 0.49M (10.5K × 47)          & 99.38        & 96.91           & 94.50          & 89.78          & 93.35          & 94.78         \\
\midrule
SynFace~\cite{qiu2021synface}         & ICCV 21      & GAN        & 0.5M (10K × 50)             & 91.93        & 75.03           & 61.63          & 70.43          & 74.73          & 74.75         \\
SFace~\cite{boutros2022sface}         & IJCB 22      & GAN        & 0.6M (10K × 60)             & 91.87        & 73.86           & 71.68          & 73.20          & 77.93          & 77.71         \\
DigiFace~\cite{bae2023digiface}       & WACV 23      & Rendering  & 0.5M (10K × 50)             & 95.40        & 87.40           & 76.97          & 78.87          & 78.62          & 83.45         \\
IDNet~\cite{kolf2023identity}         & CVPRW 23      & GAN        & 0.5M (10K × 50)             & 84.83        & 70.43           & 63.58          & 67.35          & 71.50          & 71.54         \\
DCFace~\cite{kim2023dcface}           & CVPR 23      & DM         & 0.5M (10K × 50)             & 98.55        & 85.33           & 89.70          & 82.62          & 91.60          & 89.56         \\
IDiff-Face~\cite{boutros2023idiff}    & ICCV 23      & DM         & 0.5M (10K × 50)             & 98.00        & 85.47           & 86.43          & 80.45          & 90.65          & 88.20         \\
ExFaceGAN~\cite{boutros2023exfacegan} & IJCB 23      & GAN        & 0.5M (10K × 50)             & 93.50        & 73.84           & 78.92          & 71.60          & 82.98          & 80.17         \\
USynthFace~\cite{boutros2023unsupervised} & FG 23    & GAN & 0.4M (400K × 1) & 92.23 & 78.56 & 71.62          & 72.03 & 77.05          & 78.30 \\
SFace2~\cite{boutros2024sface2}       & TBIOM 24      & GAN        & 0.6M (10K × 60)             & 94.62        & 76.24           & 74.37          & 81.57          & 72.18          & 79.80         \\
Arc2Face~\cite{papantoniou2024arc2face} & ECCV 24    & DM         & 0.5M (10K × 50)             & 98.81        & 91.87           & 90.18          & 85.16          & 92.63          & 91.73         \\
ID3~\cite{li2024id}                   & NeurIPS 24   & DM         & 0.5M (10K × 50)             & 97.68        & 86.84           & 91.00          & 82.77          & 90.73          & 89.80         \\
CemiFace~\cite{sun2024cemiface}       & NeurIPS 24   & DM         & 0.5M (10K × 50)             & 99.03        & 91.06           & 91.33          & 87.65       & 92.42          & 92.30         \\
UIFace~\cite{lin2025uiface}           & ICLR 25      & DM         & 0.5M (10K × 50)             & 99.27 & \underline{94.29} & 90.95      & \textbf{89.58} & 92.25      & 93.27 \\
Vec2Face~\cite{wu2025vec2face}         & ICLR 25  & GAN & 0.5M (10K × 50) & 98.87 & 88.97 & \textbf{93.12} & 85.47 & \textbf{93.57} & 92.00 \\
HyperFace~\cite{shahreza2025hyperface} & ICLR 25  & DM  & 0.5M (10K × 50) & 98.50 & 88.83 & 86.53          & 84.23 & 89.40          & 89.50 \\
MorphFace~\cite{mi2025data}           & CVPR 25      & DM         & 0.5M (10K × 50)             & 99.25        & 94.11           & 91.80          & 88.73          & 92.73          & \underline{93.32}         \\
NegFaceDiff~\cite{caldeira2025negfacediff} & ICCVW 25 & DM & 0.5M (10K × 50) & 98.98 & 91.67 & 90.02       & 88.82 & 91.65          & 92.23 \\
IDPerturb~\cite{boutros2026idperturb} * & CVPR 26    & DM         & 0.5M (10K × 50)             & \underline{99.32}        & 94.16           & 91.30          & 88.53          & 92.38          & 93.14         \\ % update using IDPerturb's open-souce code, using recommended $lb=0.6$
\rowcolor{gray!15}\textbf{SteerFace}                              & (Ours)       & DM         & 0.5M (10K × 50)             & \textbf{99.40} & \textbf{94.67} & \underline{92.73} & \underline{88.85} & \underline{92.95} & \textbf{93.72} \\
\midrule
DigiFace~\cite{bae2023digiface}       & WACV 23      & Rendering  & 1.2M (10K × 72, 100K × 5)   & 96.17        & 89.81           & 81.10          & 82.23          & 82.55          & 86.37         \\
DCFace~\cite{kim2023dcface}           & CVPR 23      & DM         & 1.2M (20K × 50, 40K × 5)    & 98.58        & 88.61           & 90.97          & 85.07          & 92.82          & 91.21         \\
Arc2Face~\cite{papantoniou2024arc2face} & ECCV 24    & DM         & 1.2M (20K × 50, 40K × 5)    & 98.92        & 94.58           & 92.45          & 86.45          & 93.33          & 93.15         \\
UIFace~\cite{lin2025uiface} *         & ICLR 25      & DM         & 1.2M (24K × 50)             & \underline{99.27} & 95.13 & 92.51 & \underline{90.63} & 93.17 & 94.14 \\
MorphFace~\cite{mi2025data}           & CVPR 25      & DM         & 1.2M (24K × 50)             & \textbf{99.35} & 94.77         & \textbf{93.27} & 90.07          & \underline{93.40} & \underline{94.17} \\
% IDPerturb~\cite{boutros2026idperturb} * & CVPR 26    & DM         & 1.2M (24K × 50)             & 99.33        & 94.92           & 92.16          & 89.27          & 92.92          & 93.72         \\ % update using recommended $lb=0.6$
% IDPerturb~\cite{boutros2026idperturb} * & CVPR 26    & DM         & 1.2M (24K × 50)             & \textbf{99.35}        & 95.03           & 92.21          & 89.42          & 93.03          & 93.81         \\
IDPerturb~\cite{boutros2026idperturb} * & CVPR 26    & DM         & 1.2M (24K × 50)             & \textbf{99.35}        & \underline{95.21}           & 92.47          & 89.72          & 93.38          & 94.03         \\ % update using their open-souce code, using recommended $lb=0.6$
\rowcolor{gray!15} \textbf{SteerFace}                              & (Ours)       & DM         & 1.2M (24K × 50)             & \textbf{99.35} & \textbf{95.27} & \underline{93.12}        & \textbf{90.66} & \textbf{93.43} & \textbf{94.37} \\
\bottomrule
\end{tabular}
\end{table*}

\section{Experiments}
\label{sec:exp}

\subsection{Experimental Setup}
\label{subsec:exp-setup}

\begin{figure}[tbp]
    \centering
    \includegraphics[width=\linewidth]{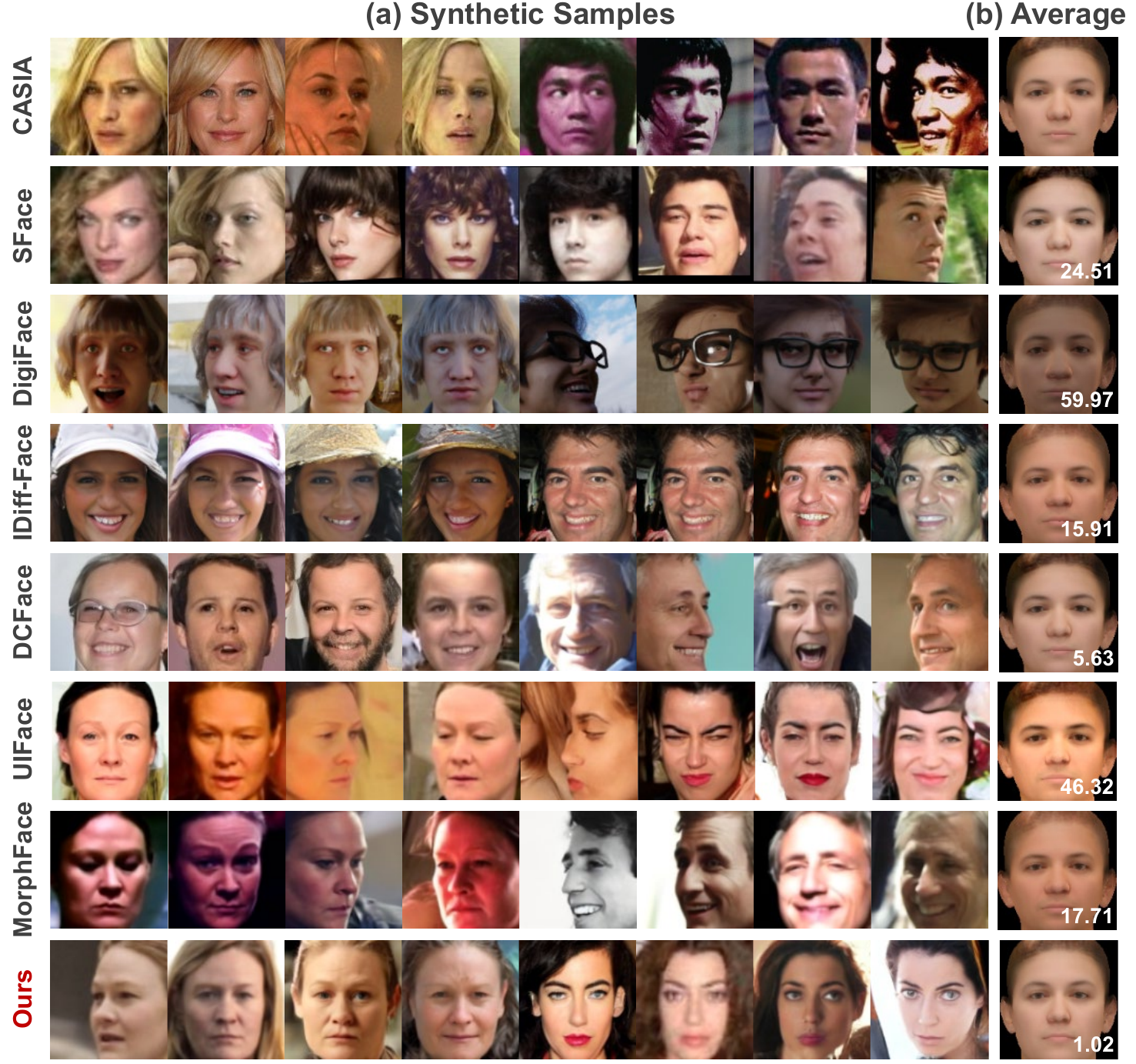}
    \caption{Samples and average faces of real data~\cite{yi2014learning}, SteerFace and SOTAs~\cite{boutros2022sface,bae2023digiface,boutros2023idiff,kim2023dcface,lin2025uiface,mi2025data}. (a) Synthetic samples from two virtual subjects. (b) 3DMM renderings of dataset-wise average faces. Numbers in the corner indicate the synthetic-real distribution gap measured by MSE, where lower is better. SteerFace achieves the smallest gap among all SOTAs.
    % for internal review; its stupid
    % NOTE: Face images are from CASIA, SFace, DigiFace, IDiff-Face, DCFace, UIFace and MorphFace~\cite{boutros2022sface,bae2023digiface,boutros2023idiff,kim2023dcface,lin2025uiface,mi2025data}.
    }
    \label{fig:comp-sota}
    
\end{figure}

\subsubsection{Generator} 
We instantiate the generator $\mathcal{G}$ based on the open-source code of IDiff-Face~\cite{boutros2023idiff}, and train it on CASIA-WebFace~\cite{yi2014learning}. The dataset contains 490K face images from 10,575 identities with varying image quality. We use an off-the-shelf ElasticFace~\cite{boutros2022elastic} identity extractor $\mathcal{F}$, and the encoder $\mathcal{\phi}_e$ and decoder $\mathcal{\phi}_d$ from Stable Diffusion~\cite{rombach2022high}. We train $\mathcal{G}$ for 250K steps using the Adam optimizer~\cite{kingma2014adam}, with an initial learning rate of $10^{-4}$ and a total batch size of 512. We set $w_{\mathrm{id}}=10$, $w_{\mathrm{reg}}=1$, and $k=10$.

\subsubsection{Synthetic face generation} 
We use the virtual identity set $V$ of 10K/24K reference face images publicly released by UIFace~\cite{lin2025uiface}, which are generated by an unconditional DM~\cite{kim2023dcface}. For each identity, we synthesize 50 images at a resolution of $128^2$, yielding two synthetic datasets of 0.5M/1.2M images, respectively.

\subsubsection{Downstream FR} 
We adopt IR-50~\cite{he2016deep} as the downstream FR model. It takes synthetic images resized to $112^2$ as input, and is trained for 32 epochs using the ArcFace~\cite{deng2019arcface} loss, with a batch size of 256 and a learning rate of 0.1, divided by 10 at epochs 22 and 28. We evaluate the trained model on 5 widely used benchmarks: LFW~\cite{lfwtechupdate}, CFP-FP~\cite{sengupta2016frontal}, AgeDB~\cite{moschoglou2017agedb}, CPLFW~\cite{zheng2018cross}, and CALFW~\cite{zheng2017cross}.

\subsection{Comparison with SOTAs}
\label{subsec:exp-comp-sota}

\subsubsection{FR performance}
We compare SteerFace with a real-data baseline, namely training on CASIA~\cite{yi2014learning}, as well as with 18 SOTAs.
% SynFace~\cite{qiu2021synface}, SFace~\cite{boutros2022sface}, DigiFace~\cite{bae2023digiface}, IDNet~\cite{kolf2023identity}, DCFace~\cite{kim2023dcface}, IDiff-Face~\cite{boutros2023idiff}, ExFaceGAN~\cite{boutros2023exfacegan}, SFace2~\cite{boutros2024sface2}, Arc2Face~\cite{papantoniou2024arc2face}, ID3~\cite{li2024id}, CemiFace~\cite{sun2024cemiface}, UIFace~\cite{lin2025uiface}, MorphFace~\cite{mi2025data}, and IDPerturb~\cite{boutros2026idperturb}. 
See~\cref{subsec:rw-ipfs-sota} for their discussion. Notably, IDPerturb~\cite{boutros2026idperturb} works with different generation pipelines, and its reported results use a pretrained UIFace~\cite{lin2025uiface} model. This flexibility is useful, but as UIFace is already more advantageous than the commonly used base framework IDiff-Face~\cite{boutros2023idiff}, it gives IDPerturb a stronger starting point. To align the pipelines and isolate the effect of embedding perturbation, we compare two settings. In~\cref{tab:exp-fr-performance}, we reproduce IDPerturb on IDiff-Face using its official code\footnote{\url{https://github.com/fdbtrs/IDperturb}} and recommended settings. In~\cref{tab:exp-generality}, we report its original results alongside SteerFace integrated with UIFace. We note that SteerFace achieves higher average FR accuracy than both UIFace-based and IDiff-Face-based IDPerturb. % Why we reproduce IDPerturb should be made clear.

% --- camera-ready ---
% Notably, IDPerturb uses a pretrained UIFace model, giving it a stronger base than other IDiff-Face-based methods. For a fair comparison, we reproduce it on IDiff-Face to isolate its standalone effect in~\cref{tab:exp-fr-performance}, and compare IDPerturb and SteerFace on UIFace in~\cref{tab:exp-generality}. Generator training datasets may also vary across methods; see the supplementary material for details.
% ------

% Notably, IDPerturb is by default instantiated from UIFace and thus enjoys a more advantageous setting than other SOTAs. Here we isolate its standalone effect, while evaluating their integration later in~\cref{tab:exp-generality}. Results are shown in~\cref{tab:exp-fr-performance}.

% --- camera-ready ---
% We observe that SteerFace outperforms all SOTAs on most benchmarks as well as on average. It surpasses the best SOTA by 0.4/0.2\% on average at the 0.5M/1.2M scales, and by 0.38\% and 0.93\% on CFP-FP and AgeDB, respectively. It also reduces the synthetic-real gap to 1.06\%. Moreover, SteerFace at 0.5M already surpasses half of the SOTAs at 1.2M, demonstrating its strong effectiveness.
% ------

SteerFace achieves the highest average FR accuracy at both scales, outperforming the best competing method by 0.4\% and 0.2\% at 0.5M and 1.2M, and by 0.38\% on CFP-FP, respectively. It also reduces the synthetic-to-real gap to 1.06\%. Moreover, SteerFace at 0.5M already surpasses half of the compared methods at 1.2M.

% % Notably, SteerFace is instantiated from IDiff-Face with the proposed perturbation incorporated. The gap between the two suggests that the proposed perturbation brings substantial gains, which we attribute to its effect of debiasing visual tendency. A more aligned comparison is later presented in~\cref{subsec:exp-perturb}.

\subsubsection{Visualization}
We visualize samples from SOTAs with publicly available datasets, and compare them with CASIA and SteerFace in~\cref{fig:comp-sota}(a). We observe that SFace and DigiFace, as earlier GAN- or rendering-based methods, fall short in either identity retention or photorealism. IDiff-Face is among the foundational works that established identity adherence in DM-based generation, but shows limited diversity, with samples from the same identity appearing nearly identical. Later methods improve diversity, yet still exhibit noticeable visual tendency: DCFace generates a large proportion of smiling faces, while UIFace and MorphFace show unrealistic color tones. By contrast, SteerFace produces samples that are both diverse and more visually unbiased. Further see~\cref{subsec:exp-quality}.

\subsection{Synthetic Data Quality}
\label{subsec:exp-quality}

We evaluate synthetic data quality by how closely it mimics real data, based on three criteria: \textit{identity adherence}, \textit{data diversity}, and, as newly focused in this work, overall \textit{visual tendency}.

\begin{figure}[tbp]
    \centering
    \includegraphics[width=0.85\linewidth]{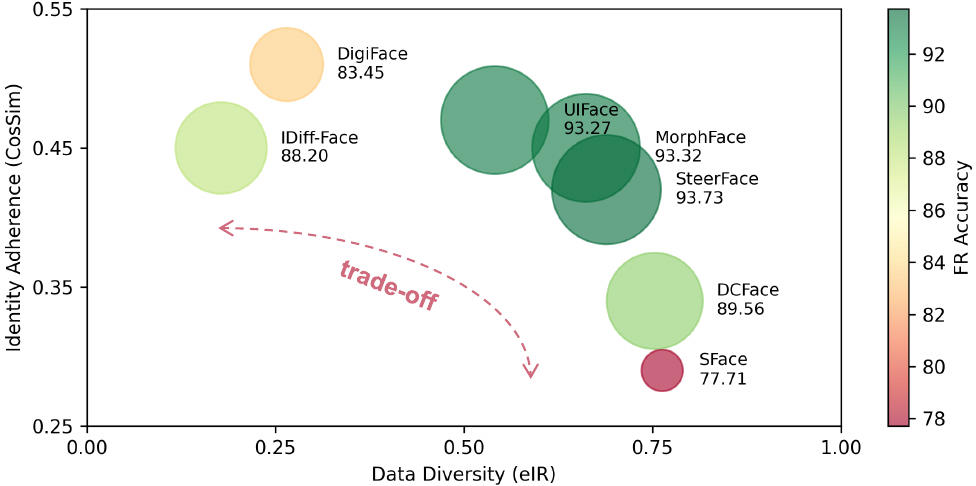}
    \caption{Comparison of SteerFace and SOTAs in terms of identity adherence and data diversity. Circle color and size indicate downstream FR accuracy. SteerFace achieves a favorable balance and delivers superior FR performance.}
    \label{fig:sim-eir}
\end{figure}

\subsubsection{Identity adherence vs. Data diversity}
Quantitatively, we measure identity adherence using the cosine similarity $\langle \mathbf{c}_1,\mathbf{c}_2 \rangle$ between intra-person embeddings extracted by an off-the-shelf FR model~\cite{boutros2022elastic}, where higher similarity indicates stronger adherence. For diversity, we use extended improved recall (eIR)~\cite{kim2023dcface}, which measures the sparsity of the style-feature manifold, with larger values indicating better diversity. \Cref{fig:sim-eir} plots prior SOTAs and SteerFace in the similarity-eIR space. Overall, a clear trade-off is observed, with most SOTAs favoring either adherence or diversity. SteerFace lies near the top-right frontier, indicating a better balance between the two. In particular, SteerFace is implemented based on IDiff-Face yet achieves substantially higher diversity. While SteerFace does not explicitly encourage diversity, it promotes it naturally by mitigating visual tendency, thereby allowing generation to vary more freely across the overall data distribution. % Nonetheless, a salient real-synthetic gap remains, calling for future work.

\subsubsection{Visual tendency}
Visual tendency reflects bias in the data distribution. For each SOTA, we use the same 3DMM~\cite{feng2021learning} as in~\cref{fig:paradigm} to extract the dataset-level average of visual attributes and render it into an ``average face''. The results are shown in~\cref{fig:comp-sota}(b), with the mean squared error (MSE) to the real-world ground truth marked in the corner. We observe that all SOTAs exhibit visual tendency to varying degrees, most significantly in color tone. By contrast, the average face of SteerFace is much closer to the ground truth, as indicated by its lowest MSE. This suggests that SteerFace effectively mitigates visual tendency, as intended.

\subsection{Effect of Residue Perturbation}
\label{subsec:exp-perturb}

\begin{table}[tbp]

\small
\centering
\caption{Comparison of different intensity allocation strategies in terms of theoretical identity retention and regularization, and experimental similarity and FR performance.}
\label{tab:exp-perturb}

\begin{tabular}{llcccc}
\toprule
    & \textbf{Settings}    & $\mathbb{E}[\cos(\theta)]$ & $\mathbb{E}[\sin^2(\theta)]$ & \textbf{Cossim} & \textbf{FR Avg.} \\
\midrule
\textbf{(a)} & Baseline    & /                 & /                                    & 0.461       & 91.84      \\
\midrule
\multirow{4}{*}{\textbf{(b)}} & $U(0,0.75)$   & 0.784              & 0.350                                 & 0.424       & 92.68      \\
                     & \gcell $U(0,1)$      & \gcell 0.637              & \gcell 0.500                                  & \gcell 0.384       & \gcell 93.21      \\
                     & $\mathrm{Beta}(1.6,1)$ & 0.520              & 0.639                                 & 0.338       & 93.40      \\
                     & $\mathrm{Beta}(2,1)$   & 0.463              & 0.703                                 & 0.244       & /          \\
\midrule
\multirow{2}{*}{\textbf{(c)}}   & $w_{\mathrm{reg}}=0.1$       & 0.646               & 0.526                                  & 0.394       & 93.66     \\
                                & \gcell$w_{\mathrm{reg}}=1$       & \gcell 0.627               & \gcell 0.565                                  & \gcell 0.387       & \gcell 93.72      \\
\bottomrule
\end{tabular}
\end{table}
\begin{figure}[tbp]
    \centering
    \includegraphics[width=\linewidth]{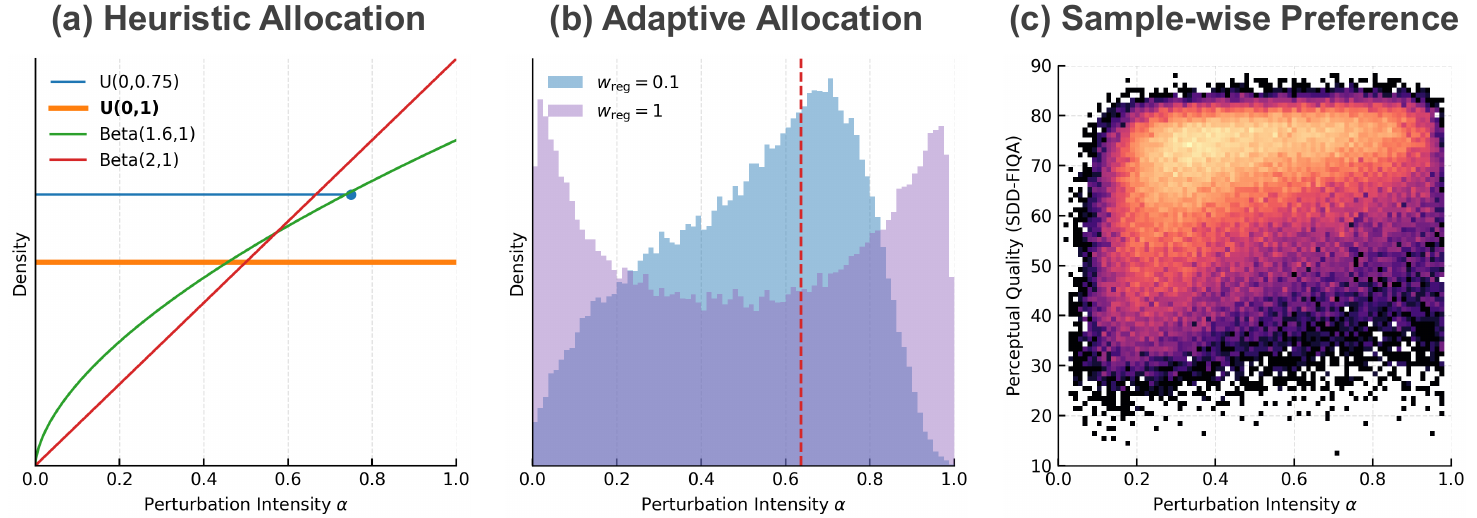}
    \caption{Allocation of perturbation intensity. (a) Four heuristic distributions for intensity allocation with growing strength. (b) Adaptively learned allocations under different $w_{\mathrm{reg}}$ differ markedly. (c) Sample-wise preference, where higher-quality samples tend to receive stronger perturbation.}
    \label{fig:perturb_dist}
\end{figure}

Regarding residue perturbation, we investigate three key questions: \textit{Is it effective? How does its effect relate to intensity allocation? Does it exhibit sample-wise preference adaptively?} We therefore compare different heuristic and adaptive allocations in~\cref{tab:exp-perturb}. For each setting, we report the \textit{theoretical} statistics of identity retention and regularization as per~\cref{thm:regularizer}, \ie, $\mathbb{E}[\cos(\theta)]$ and $\mathbb{E}[\sin^2(\theta)]$, as well as the \textit{experimental} cosine similarity and FR performance.

\subsubsection{With vs. Without perturbation}
We compare a baseline generator, trained without perturbation, with the default setting $w_{\mathrm{reg}}=1$. In~\cref{tab:exp-perturb}(a), the baseline shows better identity adherence as identity is not attenuated. However, the default setting outperforms 1.88\% in FR performance, which supports the effectiveness of perturbation.

\subsubsection{Heuristic intensity allocation}
We allocate intensity $\alpha$ from 4 heuristic distributions: $U(0,0.75)$, $U(0,1)$ (the default in~\cref{subsec:method-steer}), $\mathrm{Beta}(1.6,1)$, and $\mathrm{Beta}(2,1)$, shown in~\cref{fig:perturb_dist}(a). As shown in~\cref{tab:exp-perturb}(b), they trade off identity retention and regularization as perturbation strength increases. The corresponding cosine similarity and FR performance vary consistently, indicating that the theoretical trends are also reflected empirically. Overall, downstream FR favors stronger perturbation, with larger $\mathbb{E}[\sin^2(\theta)]$ yielding higher average accuracy, provided that identity retention remains sufficient. Under $\mathrm{Beta}(2,1)$, however, the intra-person similarity becomes too low for training an FR model effectively.

% In~\cref{subsec:method-steer}, perturbation intensity is heuristically sampled as $\alpha \sim U(0,1)$. We further compare three alternatives: $U(0,0.75)$, $\mathrm{Beta}(1.6,1)$, and $\mathrm{Beta}(2,1)$, whose distributions are shown in {\color{blue}Fig.~X(a)}. By~\cref{thm:regularizer}, their effects on identity retention and regularization should be governed by $\mathbb{E}[\cos(\theta)]$ and $\mathbb{E}[\sin^2(\theta)]$, respectively. Results in~\cref{tab:exp-perturb}(b) are consistent with this analysis: the cosine similarity of synthetic faces decreases as $\mathbb{E}[\cos(\theta)]$ becomes smaller. We also observe that downstream FR generally favors stronger perturbation, with larger $\mathbb{E}[\sin^2(\theta)]$ yielding higher average FR accuracy. This holds only as long as identity retention remains sufficient; under $\mathrm{Beta}(2,1)$, the intra-person similarity becomes too low and thus hinders FR training.

\subsubsection{Adaptive intensity allocation}
% In~\cref{subsec:method-adaptive}, the intensity distribution is adaptively learned based on the heuristic $U(0,1)$ under the weights $w_{\mathrm{id}}$ and $w_{\mathrm{reg}}$. Since identity retention $\mathcal{L}_{\mathrm{id}}$ acts as a consistency constraint, $w_{\mathrm{id}}$ is empirically less sensitive. We therefore compare $w_{\mathrm{reg}}$ at 1 (default) and 0.1. Interestingly, \cref{fig:perturb_dist}(b) shows that their learned distributions differ markedly: with $w_{\mathrm{reg}}=0.1$, the distribution favors identity retention and thus concentrates around $\mathbb{E}_{\alpha\sim U(0,1)}[\cos\theta]\approx 0.637$ (red line); with $w_{\mathrm{reg}}=1$, more mass shifts toward the high-intensity tail to strengthen regularization, while the low-intensity tail is also reinforced to preserve identity.
In~\cref{subsec:method-adaptive}, the intensity distribution is learned under the weights $w_{\mathrm{id}}$ and $w_{\mathrm{reg}}$, initialized from the heuristic $U(0,1)$. Since $\mathcal{L}{\mathrm{id}}$ acts as a consistency constraint, $w{\mathrm{id}}$ is empirically less sensitive; we therefore compare $w_{\mathrm{reg}}=1$ and $0.1$. As shown in~\cref{fig:perturb_dist}(b), their learned distributions differ markedly. With $w_{\mathrm{reg}}=0.1$, the allocation concentrates at moderate intensities and achieves expected identity retention close to $U(0,1)$ ($0.646$ vs.~$0.637$). With $w_{\mathrm{reg}}=1$, more mass shifts toward the high-intensity tail to strengthen regularization, while the low-intensity tail helps maintain comparable identity retention ($0.627$).

Despite their distinct shapes, \cref{tab:exp-perturb} shows that the two distributions yield similar $\mathbb{E}[\sin^2(\theta)]$ and close FR performance. This supports the analysis in~\cref{thm:regularizer}, showing not only that FR performance is governed primarily by regularization strength, but also that it is less sensitive to the exact distributional shape. We further find that both adaptive distributions outperform the heuristic baseline $U(0,1)$, as they maintain comparable $\mathbb{E}[\cos(\theta)]$ while achieving larger $\mathbb{E}[\sin^2(\theta)]$. This explains their FR gains and validates the effectiveness of adaptive perturbation.

% In~\cref{subsec:method-adaptive}, the perturbation distribution is adaptively refined from the heuristic $U(0,1)$, where identity retention and regularization objectives $\mathcal{L}_{\mathrm{id}}, \mathcal{L}_{\mathrm{reg}}$ are weighted by $w_{\mathrm{id}}, w_{\mathrm{reg}}$, respectively. Here, $w_{\mathrm{id}}$ is less sensitive in practice, as $\mathcal{L}_{\mathrm{id}}$ serves as a consistency constraint. We therefore compare two choices of $w_{\mathrm{reg}}$, namely 1 (default) and 0.1. As shown in {\color{blue}Fig.~X(b)}, the learned distributions differ markedly. When $w_{\mathrm{reg}}=0.1$, the distribution is centered around $\alpha=2/\pi$, favoring identity retention. When $w_{\mathrm{reg}}=1$, more probability mass is shifted toward the high-intensity tail to strengthen regularization, while the lower-intensity tail is simultaneously strengthened to compensate for $\mathbb{E}[\cos(\theta)]$. As a result, the two learned distributions differ substantially in shape, yet end up with very similar $\mathbb{E}[\sin^2(\theta)]$ and, interestingly, very similar downstream FR performance. This supports our analysis that FR performance is governed primarily by perturbation strength as measured by $\mathbb{E}[\sin^2(\theta)]$, and is less sensitive to the exact distributional shape. We further find that both adaptive distributions outperform $U(0,1)$, as they remain comparable in $\mathbb{E}[\cos(\theta)]$ while achieving larger $\mathbb{E}[\sin^2(\theta)]$. This explains their better FR accuracy and validates the effectiveness of adaptive perturbation.

\subsubsection{Sample-wise preference}
We empirically find that the learned perturbation intensity correlates with per-image perceptual quality. Specifically, we use SDD-FIQA~\cite{ou2021sdd} to assign quality scores to synthetic samples, and plot their correlation with the adaptively allocated intensity. As shown in~\cref{fig:perturb_dist}(c), higher-quality samples tend to be assigned larger perturbations. We conjecture that such images contain richer high-frequency details, making their residue more biased and therefore in greater need of perturbation. We leave this interesting phenomenon for future investigation.

% We empirically find that the learned perturbation intensity correlates with per-image perceptual quality. Specifically, we use SDD-FIQA~\cite{ou2021sdd} to assign quality scores to synthetic samples, and plot their correlation with the adaptively allocated intensity. As shown in {\color{blue}Fig.~X(c)}, higher-quality samples tend to be associated with moderate-to-large perturbation intensities, rather than very small ones. This suggests that the adaptive allocator preferentially assigns stronger perturbation to images with richer visual details. We conjecture that such images contain more biased residue, and therefore benefit from stronger regularization.

\subsection{Generalizability}
\label{subsec:exp-general}

\begin{table}[tbp]

\small
\centering
\caption{Generalizability of SteerFace, evaluated across different training datasets and generation pipelines.}
\label{tab:exp-generality}

\begin{tabular}{llcccc}
\toprule
\textbf{Dataset}       & \textbf{Setting} & \textbf{AgeDB} & \textbf{CPLFW} & \textbf{CALFW} & \textbf{Avg.} \\
\midrule
\multirow{2}{*}{FFHQ}  & Baseline         & 78.53          & 75.98          & 86.53          & 82.16         \\
                       & \gcell$U(0,1)$         & \gcell\textbf{84.08}          & \gcell\textbf{77.48}          & \gcell\textbf{88.98}          & \gcell\textbf{84.54}         \\
\midrule
\multirow{2}{*}{MS1M}  & Baseline         & 89.68          & 86.65          & 91.58          & 91.83         \\
                       & \gcell$U(0,1)$         & \gcell\textbf{90.78}          & \gcell\textbf{86.88}          & \gcell\textbf{92.12}          & \gcell\textbf{92.25}         \\
\midrule
\multirow{3}{*}{CASIA} & UI + Baseline    & 92.85          & 87.92          & 93.25          & 93.36         \\
                       & UI + ~\cite{boutros2026idperturb}   & \textbf{93.61}          & \underline{88.37}          & \underline{93.50}          & \underline{93.62}         \\                    
                       & \gcell UI + $U(0,1)$    & \gcell\underline{92.95}          & \gcell\textbf{89.05}          & \gcell\textbf{93.52}          & \gcell\textbf{93.88}         \\
\bottomrule
\end{tabular}
\end{table}

We further apply SteerFace to two alternative training datasets with substantially different scales: FFHQ~\cite{karras2019style} with 52K samples and MS-Celeb-1M~\cite{guo2016ms} with 5.2M samples. We also incorporate it into the alternative pipeline of UIFace~\cite{lin2025uiface}. Notably, since IDPerturb~\cite{boutros2026idperturb} is by default built on UIFace, we also include it in the comparison. We instantiate SteerFace with the heuristic $U(0,1)$ for alignment with IDPerturb; see~\cref{subsec:exp-perturb-sota}. Results are reported in~\cref{tab:exp-generality}, with LFW and CFP-FP deferred to the supplementary material.

Overall, SteerFace improves over the baseline in all settings: by 2.38\% on FFHQ, 0.42\% on MS1M, and 0.52\% when integrated into UIFace, demonstrating good generalizability. It also outperforms IDPerturb under the latter's recommended setting. Interestingly, the gains are larger on FFHQ and CASIA (\cref{tab:exp-perturb}), suggesting that SteerFace may be particularly beneficial when the training dataset is small and thus more prone to bias.

\subsection{Alternative Perturbation Strategies}
\label{subsec:exp-perturb-sota}

We compare SteerFace with two closely related lines of work that also perturb embeddings, as discussed in~\cref{subsec:method-steer}: context dropout~\cite{ho2022classifier} and inference-time perturbation~\cite{li2024id,sun2024cemiface,boutros2026idperturb,wu2025vec2face}. SteerFace is fundamentally different from both and outperforms them.

\begin{figure}[tbp]
    \centering
    \includegraphics[width=\linewidth]{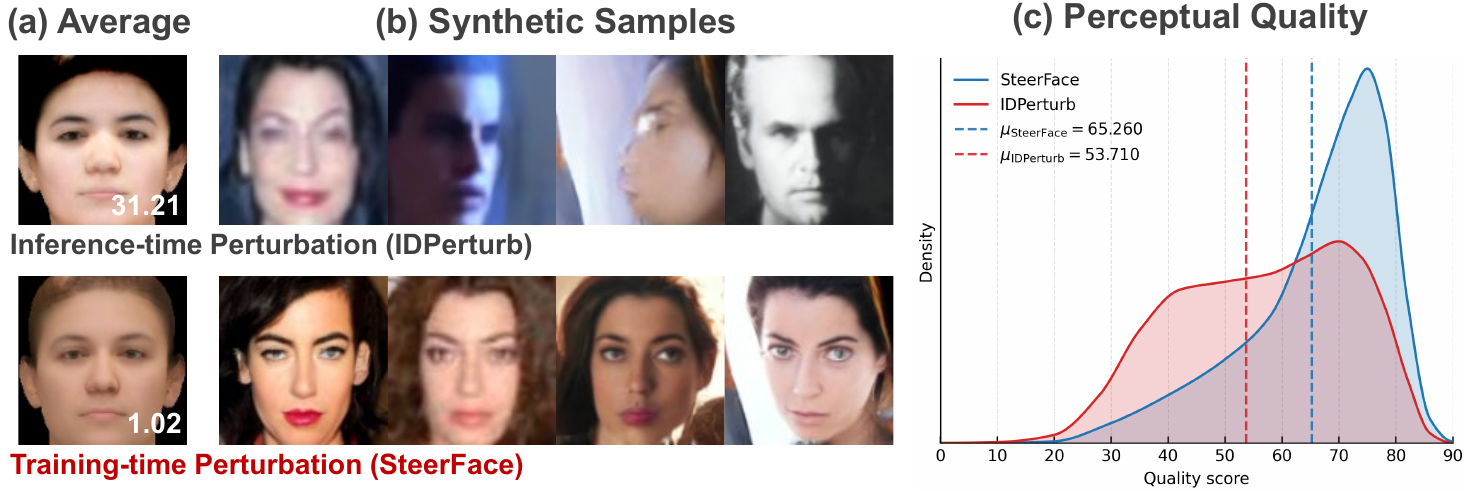}
    \caption{Inference- \textit{vs.} training-time perturbation, instantiated as IDPerturb~\cite{boutros2026idperturb} \textit{vs.} SteerFace. (a) Rendered average faces; IDPerturb remains distributionally biased. (b) Synthetic samples; SteerFace produces higher-quality images. (c) Sample-quality statistics; SteerFace is 11.5\% higher in SDD-FIQA.
    % for internal review
    % NOTE: Face images are generated via~\cite{boutros2026idperturb} based on virtual synthetic subjects~\cite{lin2025uiface}.
    }
    \label{fig:comp-perturb-fiqa}
\end{figure}

\subsubsection{Context dropout}
% --- camera-ready---
% It regularizes model learning by setting embeddings to zero with a given probability. However, visual tendency can still be learned from the unpadded residues, while identity retention is compromised because dropout perturbs identity and residue components indiscriminately. Experimentally, dropout yields at best 92.32\% FR performance, \ie, 1.4\% below SteerFace.
% ------
It regularizes model learning by setting embeddings to zero with a given probability. Specifically, we use CFG-style dropout~\cite{ho2022classifier}, which zeros the entire context and should be distinguished from the dimension-wise dropout (CPD) in IDiff-Face~\cite{boutros2023idiff}. We retain 25\% CPD here and across our experiments. Experimentally, we find that dropout yields at best 92.32\% FR performance, \ie, 1.4\% below SteerFace. This may be because visual tendency can still be learned from the retained contexts, while dropout compromises identity retention by removing identity and residue components indiscriminately.

\subsubsection{Inference-time perturbation} 
% --- camera-ready ---
% It improves data diversity by perturbing embeddings within a local neighborhood on $\mathbb{S}^{d-1}$, similarly to~\cref{fig:tendency}(c), to introduce more feature variations. IDPerturb~\cite{boutros2026idperturb} is closest to our method, as it also applies spherical rotation (\cref{eq:rotation}), but at inference time. However, such perturbation cannot mitigate visual tendency, as shown in~\cref{fig:comp-perturb-fiqa}(a), since the bias is inherited during training and only manifests differently at inference time.
% Moreover, inference-time perturbation may introduce artifacts by disrupting embedding semantics. In~\cref{fig:comp-perturb-fiqa}(b), IDPerturb shows more failure cases than SteerFace. We further assess perceptual quality using SDD-FIQA~\cite{ou2021sdd} and plot the score distributions in~\cref{fig:comp-perturb-fiqa}(c). IDPerturb's left-shifted distribution indicates statistically degraded sample quality due to such artifacts. For FR performance, since IDPerturb employs a uniform perturbation~\cite{boutros2026idperturb}, we instantiate SteerFace with the same heuristic $U(0,1)$ for a matched comparison. \Cref{tab:exp-generality} shows that SteerFace outperforms IDPerturb, suggesting the standalone advantage of training-time perturbation.
% ------
It improves data diversity by perturbing embeddings within a local neighborhood on $\mathbb{S}^{d-1}$, similarly to~\cref{fig:tendency}(c), to introduce more feature variations. IDPerturb~\cite{boutros2026idperturb} is closest to our method, as it also applies spherical rotation (\cref{eq:rotation}), but at inference time. % However, visual tendency remains under such perturbation, as shown in~\cref{fig:comp-perturb-fiqa}(a). This is consistent with the bias being learned during training and remaining in the pretrained generator.

Inference-time perturbation offers a convenient way to enrich diversity without retraining the generator. However, the generator has not been explicitly trained to accommodate these perturbations, which may alter both identity and visual cues in the context. The resulting changes can increase appearance variation, but may also introduce artifacts when the perturbed context is not well supported by the learned mapping. Training-time perturbation allows the generator to adapt to such changes. \Cref{fig:comp-perturb-fiqa}(b) illustrates this difference through sample generations from IDPerturb and SteerFace, where IDPerturb introduces visible artifacts. We further assess perceptual quality using SDD-FIQA~\cite{ou2021sdd} and plot the score distributions in~\cref{fig:comp-perturb-fiqa}(c), where IDPerturb's left-shifted distribution indicates lower perceptual quality on average. 

In~\cref{tab:exp-generality}, we also compare FR performance on UIFace, IDPerturb's original setting, where we use the heuristic $U(0,1)$ for SteerFace. Results show that SteerFace outperforms IDPerturb under the same UIFace pipeline, further supporting the benefit of training-time perturbation.

% Moreover, inference-time perturbation may introduce artifacts by disrupting embedding semantics. In~\cref{fig:comp-perturb-fiqa}(b), we show examples where IDPerturb introduces visible artifacts. We further assess perceptual quality using SDD-FIQA~\cite{ou2021sdd} and plot the score distributions in~\cref{fig:comp-perturb-fiqa}(c). IDPerturb's left-shifted distribution indicates lower perceptual quality on average. For FR performance, in~\cref{tab:exp-generality}, we compare IDPerturb and on UIFace, retaining IDPerturb's recommended uniform sampling settings and using the heuristic $U(0,1)$ for SteerFace to exclude gains from adaptive allocation. \Cref{tab:exp-generality} shows that SteerFace outperforms IDPerturb under the same UIFace pipeline, suggesting an advantage of training-time perturbation in this setting.
% See the supplementary material for a more detailed discussion.
\section{Conclusion}

This paper identifies visual tendency as a previously underexplored limitation of synthetic face generation for FR, beyond identity adherence and data diversity. When conditioned on identity embeddings, diffusion-based generators may absorb co-occurring residual visual cues into identity semantics, inducing a synthetic-real distribution bias. To address this issue, SteerFace perturbs identity embeddings toward random orthogonal directions on the hypersphere with adaptively allocated intensities, thereby regularizing the generator while preserving identity. Experiments show that SteerFace mitigates visual tendency, outperforms SOTAs in downstream FR, and generalizes across training datasets and generation pipelines.

% This paper identifies visual tendency as a previously underexplored limitation of synthetic face generation for FR, beyond identity adherence and data diversity. When conditioned on identity embeddings, diffusion-based generators may unintentionally absorb co-occurring residual visual cues into learned identity semantics, thereby inducing a synthetic-real distribution bias. To address this issue, this paper proposes SteerFace, a simple and efficient training framework that perturbs identity embeddings toward random orthogonal directions on the embedding hypersphere. The perturbation acts as an identity-preserving regularizer and is further combined with an adaptive intensity allocation strategy. Extensive experiments show that SteerFace effectively mitigates visual tendency, improves downstream FR over SOTAs, and generalizes well across different training datasets and generation pipelines.

\begin{acks}
This research was supported in part by the National Natural Science Foundation of China (Grant No.~62506080), in part by the China Postdoctoral Science Foundation (Grant No.~GZC20251072), and in part by Tencent. % The computations in this research were performed using the CFFF platform of Fudan University.
\end{acks}

% \newpage
% \input{rebuttal}

% \begin{acks}
% \end{acks}

\bibliographystyle{ACM-Reference-Format}
\balance
\bibliography{main}

@inproceedings{he2016deep,
  title={Deep residual learning for image recognition},
  author={He, Kaiming and Zhang, Xiangyu and Ren, Shaoqing and Sun, Jian},
  booktitle={Proceedings of the IEEE conference on computer vision and pattern recognition},
  pages={770--778},
  year={2016}
}

@InProceedings{boutros2022elastic,
    author    = {Boutros, Fadi and Damer, Naser and Kirchbuchner, Florian and Kuijper, Arjan},
    title     = {ElasticFace: Elastic Margin Loss for Deep Face Recognition},
    booktitle = {Proceedings of the IEEE/CVF Conference on Computer Vision and Pattern Recognition (CVPR) Workshops},
    month     = {June},
    year      = {2022},
    pages     = {1578-1587}
}

@inproceedings{deng2019arcface,
  title={Arcface: Additive angular margin loss for deep face recognition},
  author={Deng, Jiankang and Guo, Jia and Xue, Niannan and Zafeiriou, Stefanos},
  booktitle={Proceedings of the IEEE/CVF conference on computer vision and pattern recognition},
  pages={4690--4699},
  year={2019}
}

@inproceedings{wang2018cosface,
  title={Cosface: Large margin cosine loss for deep face recognition},
  author={Wang, Hao and Wang, Yitong and Zhou, Zheng and Ji, Xing and Gong, Dihong and Zhou, Jingchao and Li, Zhifeng and Liu, Wei},
  booktitle={Proceedings of the IEEE conference on computer vision and pattern recognition},
  pages={5265--5274},
  year={2018}
}

@inproceedings{huang2020curricularface,
  title={Curricularface: adaptive curriculum learning loss for deep face recognition},
  author={Huang, Yuge and Wang, Yuhan and Tai, Ying and Liu, Xiaoming and Shen, Pengcheng and Li, Shaoxin and Li, Jilin and Huang, Feiyue},
  booktitle={proceedings of the IEEE/CVF conference on computer vision and pattern recognition},
  pages={5901--5910},
  year={2020}
}

@inproceedings{cao2018vggface2,
  title={Vggface2: A dataset for recognising faces across pose and age},
  author={Cao, Qiong and Shen, Li and Xie, Weidi and Parkhi, Omkar M and Zisserman, Andrew},
  booktitle={2018 13th IEEE international conference on automatic face \& gesture recognition (FG 2018)},
  pages={67--74},
  year={2018},
  organization={IEEE}
}

@inproceedings{zhu2021webface260m,
  title={Webface260m: A benchmark unveiling the power of million-scale deep face recognition},
  author={Zhu, Zheng and Huang, Guan and Deng, Jiankang and Ye, Yun and Huang, Junjie and Chen, Xinze and Zhu, Jiagang and Yang, Tian and Lu, Jiwen and Du, Dalong and others},
  booktitle={Proceedings of the IEEE/CVF Conference on Computer Vision and Pattern Recognition},
  pages={10492--10502},
  year={2021}
}

@inproceedings{guo2016ms,
  title={Ms-celeb-1m: A dataset and benchmark for large-scale face recognition},
  author={Guo, Yandong and Zhang, Lei and Hu, Yuxiao and He, Xiaodong and Gao, Jianfeng},
  booktitle={Computer Vision--ECCV 2016: 14th European Conference, Amsterdam, The Netherlands, October 11-14, 2016, Proceedings, Part III 14},
  pages={87--102},
  year={2016},
  organization={Springer}
}

@inproceedings{kemelmacher2016megaface,
  title={The megaface benchmark: 1 million faces for recognition at scale},
  author={Kemelmacher-Shlizerman, Ira and Seitz, Steven M and Miller, Daniel and Brossard, Evan},
  booktitle={Proceedings of the IEEE conference on computer vision and pattern recognition},
  pages={4873--4882},
  year={2016}
}

@article{yi2014learning,
  title={Learning face representation from scratch},
  author={Yi, Dong and Lei, Zhen and Liao, Shengcai and Li, Stan Z},
  journal={arXiv preprint arXiv:1411.7923},
  year={2014}
}

@techreport{lfwtechupdate,
  author =       {Gary B. Huang Erik Learned-Miller},
  title =        {Labeled Faces in the Wild: Updates and New Reporting 
                  Procedures},
  institution =  {University of Massachusetts, Amherst},
  year =         2014,
  number =       {UM-CS-2014-003},
  month =        {May}}

@inproceedings{sengupta2016frontal,
  title={Frontal to profile face verification in the wild},
  author={Sengupta, Soumyadip and Chen, Jun-Cheng and Castillo, Carlos and Patel, Vishal M and Chellappa, Rama and Jacobs, David W},
  booktitle={2016 IEEE winter conference on applications of computer vision (WACV)},
  pages={1--9},
  year={2016},
  organization={IEEE}
}

@article{zheng2018cross,
  title={Cross-pose lfw: A database for studying cross-pose face recognition in unconstrained environments},
  author={Zheng, Tianyue and Deng, Weihong},
  journal={Beijing University of Posts and Telecommunications, Tech. Rep},
  volume={5},
  number={7},
  pages={5},
  year={2018}
}

@article{zheng2017cross,
  title={Cross-age lfw: A database for studying cross-age face recognition in unconstrained environments},
  author={Zheng, Tianyue and Deng, Weihong and Hu, Jiani},
  journal={arXiv preprint arXiv:1708.08197},
  year={2017}
}

@inproceedings{moschoglou2017agedb,
  title={Agedb: the first manually collected, in-the-wild age database},
  author={Moschoglou, Stylianos and Papaioannou, Athanasios and Sagonas, Christos and Deng, Jiankang and Kotsia, Irene and Zafeiriou, Stefanos},
  booktitle={proceedings of the IEEE conference on computer vision and pattern recognition workshops},
  pages={51--59},
  year={2017}
}

@inproceedings{karras2019style,
  title={A style-based generator architecture for generative adversarial networks},
  author={Karras, Tero and Laine, Samuli and Aila, Timo},
  booktitle={Proceedings of the IEEE/CVF conference on computer vision and pattern recognition},
  pages={4401--4410},
  year={2019}
}

@inproceedings{deng2018uv,
  title={Uv-gan: Adversarial facial uv map completion for pose-invariant face recognition},
  author={Deng, Jiankang and Cheng, Shiyang and Xue, Niannan and Zhou, Yuxiang and Zafeiriou, Stefanos},
  booktitle={Proceedings of the IEEE conference on computer vision and pattern recognition},
  pages={7093--7102},
  year={2018}
}

@inproceedings{geng20193d,
  title={3d guided fine-grained face manipulation},
  author={Geng, Zhenglin and Cao, Chen and Tulyakov, Sergey},
  booktitle={Proceedings of the IEEE/CVF conference on computer vision and pattern recognition},
  pages={9821--9830},
  year={2019}
}

@inproceedings{medin2022most,
  title={MOST-GAN: 3D morphable StyleGAN for disentangled face image manipulation},
  author={Medin, Safa C and Egger, Bernhard and Cherian, Anoop and Wang, Ye and Tenenbaum, Joshua B and Liu, Xiaoming and Marks, Tim K},
  booktitle={Proceedings of the AAAI conference on artificial intelligence},
  volume={36},
  number={2},
  pages={1962--1971},
  year={2022}
}

@inproceedings{nguyen2019hologan,
  title={Hologan: Unsupervised learning of 3d representations from natural images},
  author={Nguyen-Phuoc, Thu and Li, Chuan and Theis, Lucas and Richardt, Christian and Yang, Yong-Liang},
  booktitle={Proceedings of the IEEE/CVF International Conference on Computer Vision},
  pages={7588--7597},
  year={2019}
}

@inproceedings{piao2019semi,
  title={Semi-supervised monocular 3D face reconstruction with end-to-end shape-preserved domain transfer},
  author={Piao, Jingtan and Qian, Chen and Li, Hongsheng},
  booktitle={Proceedings of the IEEE/CVF international conference on computer vision},
  pages={9398--9407},
  year={2019}
}

@article{ho2020denoising,
  title={Denoising diffusion probabilistic models},
  author={Ho, Jonathan and Jain, Ajay and Abbeel, Pieter},
  journal={Advances in neural information processing systems},
  volume={33},
  pages={6840--6851},
  year={2020}
}

@article{song2020denoising,
  title={Denoising diffusion implicit models},
  author={Song, Jiaming and Meng, Chenlin and Ermon, Stefano},
  journal={arXiv preprint arXiv:2010.02502},
  year={2020}
}

@article{gal2022image,
  title={An image is worth one word: Personalizing text-to-image generation using textual inversion},
  author={Gal, Rinon and Alaluf, Yuval and Atzmon, Yuval and Patashnik, Or and Bermano, Amit H and Chechik, Gal and Cohen-Or, Daniel},
  journal={arXiv preprint arXiv:2208.01618},
  year={2022}
}

@inproceedings{ruiz2023dreambooth,
  title={Dreambooth: Fine tuning text-to-image diffusion models for subject-driven generation},
  author={Ruiz, Nataniel and Li, Yuanzhen and Jampani, Varun and Pritch, Yael and Rubinstein, Michael and Aberman, Kfir},
  booktitle={Proceedings of the IEEE/CVF conference on computer vision and pattern recognition},
  pages={22500--22510},
  year={2023}
}

@inproceedings{rombach2022high,
  title={High-resolution image synthesis with latent diffusion models},
  author={Rombach, Robin and Blattmann, Andreas and Lorenz, Dominik and Esser, Patrick and Ommer, Bj{\"o}rn},
  booktitle={Proceedings of the IEEE/CVF conference on computer vision and pattern recognition},
  pages={10684--10695},
  year={2022}
}

@article{gal2023encoder,
  title={Encoder-based domain tuning for fast personalization of text-to-image models},
  author={Gal, Rinon and Arar, Moab and Atzmon, Yuval and Bermano, Amit H and Chechik, Gal and Cohen-Or, Daniel},
  journal={ACM Transactions on Graphics (TOG)},
  volume={42},
  number={4},
  pages={1--13},
  year={2023},
  publisher={ACM New York, NY, USA}
}

@article{zhou2023enhancing,
  title={Enhancing detail preservation for customized text-to-image generation: A regularization-free approach},
  author={Zhou, Yufan and Zhang, Ruiyi and Sun, Tong and Xu, Jinhui},
  journal={arXiv preprint arXiv:2305.13579},
  year={2023}
}

@article{yuan2023inserting,
  title={Inserting anybody in diffusion models via celeb basis},
  author={Yuan, Ge and Cun, Xiaodong and Zhang, Yong and Li, Maomao and Qi, Chenyang and Wang, Xintao and Shan, Ying and Zheng, Huicheng},
  journal={arXiv preprint arXiv:2306.00926},
  year={2023}
}

@article{wang2024stableidentity,
  title={Stableidentity: Inserting anybody into anywhere at first sight},
  author={Wang, Qinghe and Jia, Xu and Li, Xiaomin and Li, Taiqing and Ma, Liqian and Zhuge, Yunzhi and Lu, Huchuan},
  journal={arXiv preprint arXiv:2401.15975},
  year={2024}
}

@article{xiao2024fastcomposer,
  title={Fastcomposer: Tuning-free multi-subject image generation with localized attention},
  author={Xiao, Guangxuan and Yin, Tianwei and Freeman, William T and Durand, Fr{\'e}do and Han, Song},
  journal={International Journal of Computer Vision},
  pages={1--20},
  year={2024},
  publisher={Springer}
}

@inproceedings{li2024photomaker,
  title={Photomaker: Customizing realistic human photos via stacked id embedding},
  author={Li, Zhen and Cao, Mingdeng and Wang, Xintao and Qi, Zhongang and Cheng, Ming-Ming and Shan, Ying},
  booktitle={Proceedings of the IEEE/CVF Conference on Computer Vision and Pattern Recognition},
  pages={8640--8650},
  year={2024}
}

@inproceedings{valevski2023face0,
  title={Face0: Instantaneously conditioning a text-to-image model on a face},
  author={Valevski, Dani and Lumen, Danny and Matias, Yossi and Leviathan, Yaniv},
  booktitle={SIGGRAPH Asia 2023 Conference Papers},
  pages={1--10},
  year={2023}
}

@inproceedings{peng2024portraitbooth,
  title={Portraitbooth: A versatile portrait model for fast identity-preserved personalization},
  author={Peng, Xu and Zhu, Junwei and Jiang, Boyuan and Tai, Ying and Luo, Donghao and Zhang, Jiangning and Lin, Wei and Jin, Taisong and Wang, Chengjie and Ji, Rongrong},
  booktitle={Proceedings of the IEEE/CVF Conference on Computer Vision and Pattern Recognition},
  pages={27080--27090},
  year={2024}
}

@inproceedings{ding2023diffusionrig,
  title={Diffusionrig: Learning personalized priors for facial appearance editing},
  author={Ding, Zheng and Zhang, Xuaner and Xia, Zhihao and Jebe, Lars and Tu, Zhuowen and Zhang, Xiuming},
  booktitle={Proceedings of the IEEE/CVF Conference on Computer Vision and Pattern Recognition},
  pages={12736--12746},
  year={2023}
}

@inproceedings{deandres2024frcsyn,
  title={Frcsyn challenge at cvpr 2024: Face recognition challenge in the era of synthetic data},
  author={DeAndres-Tame, Ivan and Tolosana, Ruben and Melzi, Pietro and Vera-Rodriguez, Ruben and Kim, Minchul and Rathgeb, Christian and Liu, Xiaoming and Morales, Aythami and Fierrez, Julian and Ortega-Garcia, Javier and others},
  booktitle={Proceedings of the IEEE/CVF Conference on Computer Vision and Pattern Recognition},
  pages={3173--3183},
  year={2024}
}

@article{frcsyn2025,
title = {Second FRCSyn-onGoing: Winning solutions and post-challenge analysis to improve face recognition with synthetic data},
journal = {Information Fusion},
volume = {120},
pages = {103099},
year = {2025},
issn = {1566-2535},
doi = {https://doi.org/10.1016/j.inffus.2025.103099},
url = {https://www.sciencedirect.com/science/article/pii/S1566253525001721},
author = {Ivan DeAndres-Tame and Ruben Tolosana and Pietro Melzi and Ruben Vera-Rodriguez and Minchul Kim and Christian Rathgeb and Xiaoming Liu and Luis F. Gomez and Aythami Morales and Julian Fierrez and Javier Ortega-Garcia and Zhizhou Zhong and Yuge Huang and Yuxi Mi and Shouhong Ding and Shuigeng Zhou and others}
}

@inproceedings{qiu2021synface,
  title={Synface: Face recognition with synthetic data},
  author={Qiu, Haibo and Yu, Baosheng and Gong, Dihong and Li, Zhifeng and Liu, Wei and Tao, Dacheng},
  booktitle={Proceedings of the IEEE/CVF International Conference on Computer Vision},
  pages={10880--10890},
  year={2021}
}

@inproceedings{boutros2023unsupervised,
  title={Unsupervised face recognition using unlabeled synthetic data},
  author={Boutros, Fadi and Klemt, Marcel and Fang, Meiling and Kuijper, Arjan and Damer, Naser},
  booktitle={2023 IEEE 17th International Conference on Automatic Face and Gesture Recognition (FG)},
  pages={1--8},
  year={2023},
  organization={IEEE}
}

@inproceedings{boutros2022sface,
  title={Sface: Privacy-friendly and accurate face recognition using synthetic data},
  author={Boutros, Fadi and Huber, Marco and Siebke, Patrick and Rieber, Tim and Damer, Naser},
  booktitle={2022 IEEE International Joint Conference on Biometrics (IJCB)},
  pages={1--11},
  year={2022},
  organization={IEEE}
}

@article{boutros2024sface2,
  title={Sface2: Synthetic-based face recognition with w-space identity-driven sampling},
  author={Boutros, Fadi and Huber, Marco and Luu, Anh Thi and Siebke, Patrick and Damer, Naser},
  journal={IEEE Transactions on Biometrics, Behavior, and Identity Science},
  year={2024},
  publisher={IEEE}
}

@inproceedings{kolf2023identity,
  title={Identity-driven three-player generative adversarial network for synthetic-based face recognition},
  author={Kolf, Jan Niklas and Rieber, Tim and Elliesen, Jurek and Boutros, Fadi and Kuijper, Arjan and Damer, Naser},
  booktitle={Proceedings of the IEEE/CVF Conference on Computer Vision and Pattern Recognition},
  pages={806--816},
  year={2023}
}

@inproceedings{bae2023digiface,
  title={Digiface-1m: 1 million digital face images for face recognition},
  author={Bae, Gwangbin and de La Gorce, Martin and Baltru{\v{s}}aitis, Tadas and Hewitt, Charlie and Chen, Dong and Valentin, Julien and Cipolla, Roberto and Shen, Jingjing},
  booktitle={Proceedings of the IEEE/CVF Winter Conference on Applications of Computer Vision},
  pages={3526--3535},
  year={2023}
}

@inproceedings{boutros2023exfacegan,
  title={Exfacegan: Exploring identity directions in gan’s learned latent space for synthetic identity generation},
  author={Boutros, Fadi and Klemt, Marcel and Fang, Meiling and Kuijper, Arjan and Damer, Naser},
  booktitle={2023 IEEE International Joint Conference on Biometrics (IJCB)},
  pages={1--10},
  year={2023},
  organization={IEEE}
}

@inproceedings{boutros2023idiff,
  title={Idiff-face: Synthetic-based face recognition through fizzy identity-conditioned diffusion model},
  author={Boutros, Fadi and Grebe, Jonas Henry and Kuijper, Arjan and Damer, Naser},
  booktitle={Proceedings of the IEEE/CVF International Conference on Computer Vision},
  pages={19650--19661},
  year={2023}
}

@inproceedings{kim2023dcface,
  title={Dcface: Synthetic face generation with dual condition diffusion model},
  author={Kim, Minchul and Liu, Feng and Jain, Anil and Liu, Xiaoming},
  booktitle={Proceedings of the ieee/cvf conference on computer vision and pattern recognition},
  pages={12715--12725},
  year={2023}
}

@article{sun2024cemiface,
  title={Cemiface: Center-based semi-hard synthetic face generation for face recognition},
  author={Sun, Zhonglin and Song, Siyang and Patras, Ioannis and Tzimiropoulos, Georgios},
  journal={Advances in Neural Information Processing Systems},
  volume={37},
  pages={35612--35638},
  year={2024}
}

@article{li2024id,
  title={ID3: Identity-Preserving-yet-Diversified Diffusion Models for Synthetic Face Recognition},
  author={Li, Shen and Xu, Jianqing and Wu, Jiaying and Xiong, Miao and Deng, Ailin and Ji, Jiazhen and Huang, Yuge and Feng, Wenjie and Ding, Shouhong and Hooi, Bryan},
  journal={Advances in Neural Information Processing Systems},
  volume={37},
  pages={77777--77798},
  year={2024}
}

@inproceedings{papantoniou2024arc2face,
  title={Arc2Face: A foundation model for id-consistent human faces},
  author={Papantoniou, Foivos Paraperas and Lattas, Alexandros and Moschoglou, Stylianos and Deng, Jiankang and Kainz, Bernhard and Zafeiriou, Stefanos},
  booktitle={European Conference on Computer Vision},
  pages={241--261},
  year={2024},
  organization={Springer}
}

@inproceedings{ronneberger2015u,
  title={U-net: Convolutional networks for biomedical image segmentation},
  author={Ronneberger, Olaf and Fischer, Philipp and Brox, Thomas},
  booktitle={Medical image computing and computer-assisted intervention--MICCAI 2015: 18th international conference, Munich, Germany, October 5-9, 2015, proceedings, part III 18},
  pages={234--241},
  year={2015},
  organization={Springer}
}

@article{feng2021learning,
  title={Learning an animatable detailed 3D face model from in-the-wild images},
  author={Feng, Yao and Feng, Haiwen and Black, Michael J and Bolkart, Timo},
  journal={ACM Transactions on Graphics (ToG)},
  volume={40},
  number={4},
  pages={1--13},
  year={2021},
  publisher={ACM New York, NY, USA}
}

@inproceedings{ou2021sdd,
  title={SDD-FIQA: Unsupervised face image quality assessment with similarity distribution distance},
  author={Ou, Fu-Zhao and Chen, Xingyu and Zhang, Ruixin and Huang, Yuge and Li, Shaoxin and Li, Jilin and Li, Yong and Cao, Liujuan and Wang, Yuan-Gen},
  booktitle={Proceedings of the IEEE/CVF conference on computer vision and pattern recognition},
  pages={7670--7679},
  year={2021}
}

@article{kingma2014adam,
  title={Adam: A method for stochastic optimization},
  author={Kingma, Diederik P},
  journal={arXiv preprint arXiv:1412.6980},
  year={2014}
}

@article{ho2022classifier,
  title={Classifier-free diffusion guidance},
  author={Ho, Jonathan and Salimans, Tim},
  journal={arXiv preprint arXiv:2207.12598},
  year={2022}
}

@inproceedings{xu2024chain,
  title={Chain of generation: Multi-modal gesture synthesis via cascaded conditional control},
  author={Xu, Zunnan and Zhang, Yachao and Yang, Sicheng and Li, Ronghui and Li, Xiu},
  booktitle={Proceedings of the AAAI Conference on Artificial Intelligence},
  volume={38},
  number={6},
  pages={6387--6395},
  year={2024}
}

@inproceedings{mi2024privacy,
  title={Privacy-preserving face recognition using trainable feature subtraction},
  author={Mi, Yuxi and Zhong, Zhizhou and Huang, Yuge and Ji, Jiazhen and Xu, Jianqing and Wang, Jun and Wang, Shaoming and Ding, Shouhong and Zhou, Shuigeng},
  booktitle={Proceedings of the IEEE/CVF Conference on Computer Vision and Pattern Recognition},
  pages={297--307},
  year={2024}
}

@article{zhong2024slerpface,
  title={Slerpface: face template protection via spherical linear interpolation},
  author={Zhong, Zhizhou and Mi, Yuxi and Huang, Yuge and Xu, Jianqing and Mu, Guodong and Ding, Shouhong and Zhang, Jingyun and Guo, Rizen and Wu, Yunsheng and Zhou, Shuigeng},
  journal={arXiv preprint arXiv:2407.03043},
  year={2024}
}

@inproceedings{lopez2022legal,
  title={On the legal nature of synthetic data},
  author={L{\'o}pez, Cesar Augusto Fontanillo and others},
  booktitle={NeurIPS 2022 Workshop on Synthetic Data for Empowering ML Research},
  year={2022}
}

@article{guo2024liveportrait,
  title={Liveportrait: Efficient portrait animation with stitching and retargeting control},
  author={Guo, Jianzhu and Zhang, Dingyun and Liu, Xiaoqiang and Zhong, Zhizhou and Zhang, Yuan and Wan, Pengfei and Zhang, Di},
  journal={arXiv preprint arXiv:2407.03168},
  year={2024}
}

@article{zhong2025anytalker,
  title={Anytalker: Scaling multi-person talking video generation with interactivity refinement},
  author={Zhong, Zhizhou and Ji, Yicheng and Kong, Zhe and Liu, Yiying and Wang, Jiarui and Feng, Jiasun and Liu, Lupeng and Wang, Xiangyi and Li, Yanjia and She, Yuqing and others},
  journal={arXiv preprint arXiv:2511.23475},
  year={2025}
}

@inproceedings{mi2025data,
  title={Data synthesis with diverse styles for face recognition via 3dmm-guided diffusion},
  author={Mi, Yuxi and Zhong, Zhizhou and Huang, Yuge and Yuan, Qiuyang and Zhao, Xuan and Xu, Jianqing and Ding, Shouhong and Wang, Shaoming and Guo, Rizen and Zhou, Shuigeng},
  booktitle={Proceedings of the Computer Vision and Pattern Recognition Conference},
  pages={21203--21214},
  year={2025}
}

@article{boutros2026idperturb,
  title={IDperturb: Enhancing Variation in Synthetic Face Generation via Angular Perturbation},
  author={Boutros, Fadi and Caldeira, Eduarda and Chettaoui, Tahar and Damer, Naser},
  journal={arXiv preprint arXiv:2602.18831},
  year={2026}
}

@inproceedings{lin2025uiface,
  title={UIFace: Unleashing Inherent Model Capabilities to Enhance Intra-Class Diversity in Synthetic Face Recognition},
  author={Lin, Xiao and Huang, Yuge and Xu, Jianqing and Mi, Yuxi and Zhou, Shuigeng and Ding, Shouhong},
  booktitle={International Conference on Learning Representations},
  volume={2025},
  pages={87296--87309},
  year={2025}
}

@inproceedings{xu2026smrabooth,
  title={Smrabooth: Subject and motion representation alignment for customized video generation},
  author={Xu, Xuancheng and Li, Yaning and You, Sisi and Bao, Bing-Kun},
  booktitle={Proceedings of the IEEE/CVF Conference on Computer Vision and Pattern Recognition},
  pages={16130--16141},
  year={2026}
}

@article{xu2026disco,
  title={Disco-LoRA: Disentangled Composition of Content, Style, and Motion for Multi-concept Video Customization},
  author={Xu, Xuancheng and Jia, Gengyun and Bao, Bing-Kun},
  journal={arXiv preprint arXiv:2606.26668},
  year={2026}
}

@inproceedings{xu2025clgc,
  title={Clgc: Continuous layout guidance for consistent text-to-video editing},
  author={Xu, Xuancheng and Tao, Ming and Bao, Bing-Kun},
  booktitle={2025 IEEE International Conference on Multimedia and Expo (ICME)},
  pages={1--6},
  year={2025},
  organization={IEEE}
}

@inproceedings{wu2025vec2face,
  title     = {{Vec2Face}: Scaling Face Dataset Generation with
               Loosely Constrained Vectors},
  author    = {Wu, Haiyu and Singh, Jaskirat and Tian, Sicong
               and Zheng, Liang and Bowyer, Kevin W.},
  booktitle = {The Thirteenth International Conference on
               Learning Representations},
  year      = {2025},
  url       = {https://proceedings.iclr.cc/paper_files/paper/2025/hash/3ae86071c169649bff21188c536163dc-Abstract-Conference.html}
}

@inproceedings{shahreza2025hyperface,
  title     = {{HyperFace}: Generating Synthetic Face Recognition
               Datasets by Exploring Face Embedding Hypersphere},
  author    = {Otroshi Shahreza, Hatef and Marcel, S{\'e}bastien},
  booktitle = {The Thirteenth International Conference on
               Learning Representations},
  year      = {2025},
  url       = {https://proceedings.iclr.cc/paper_files/paper/2025/hash/5edb0d52b6959d946afac7600f9f1e0c-Abstract-Conference.html}
}

@inproceedings{caldeira2025negfacediff,
  title     = {{NegFaceDiff}: The Power of Negative Context in
               Identity-Conditioned Diffusion for Synthetic Face Generation},
  author    = {Caldeira, Eduarda and Damer, Naser and Boutros, Fadi},
  booktitle = {Proceedings of the IEEE/CVF International Conference
               on Computer Vision (ICCV) Workshops},
  pages     = {5973--5983},
  month     = {October},
  year      = {2025}
}

\end{document}